\documentclass[11pt]{article}

\usepackage[final]{acl}

\usepackage{times}
\usepackage{latexsym}

\usepackage[T1]{fontenc}

\usepackage[utf8]{inputenc}

\usepackage{microtype}

\usepackage{inconsolata}

\usepackage{graphicx}
\usepackage{subcaption}
\usepackage{amsmath}
\usepackage{booktabs}
\usepackage{multirow}
\usepackage{enumitem}

\newcommand{\ConfQA}{{\sc ConfQA}}
\newcommand{\ConfRAG}{{\sc ConfRAG}}
\usepackage[table]{xcolor} 
\definecolor{lightgreen}{rgb}{0.88,1,0.88}
\usepackage{newfloat}
\DeclareFloatingEnvironment[
    fileext=los,
    listname={List of Prompts},
    name=Prompt,
    placement=tbhp,
]{prompt}

\title{{\ConfRAG}: Confidence-Guided Retrieval-Augmented Generation}

\author{
 \textbf{Yin Huang\textsuperscript{1}},
 \textbf{Yifan Ethan Xu\textsuperscript{1}},
 \textbf{Kai Sun\textsuperscript{1}},
 \textbf{Vera Yan\textsuperscript{1}},
 \textbf{Alicia Sun\textsuperscript{2}},
 \textbf{Haidar Khan\textsuperscript{1}},
 \\
 \textbf{Jimmy Nguyen\textsuperscript{1}},
 \textbf{Sean Chen\textsuperscript{1}},
 \textbf{Mohammad Kachuee\textsuperscript{1}},
 \textbf{Zhaojiang Lin\textsuperscript{1}},
 \\
 \textbf{Yue Liu\textsuperscript{1}},
 \textbf{Aaron Colak\textsuperscript{1}},
 \textbf{Anuj Kumar\textsuperscript{1}},
 \textbf{Wen-tau Yih\textsuperscript{2}},
 \textbf{Xin Luna Dong\textsuperscript{1}}
\\
 \textsuperscript{1}Meta Reality Labs,
 \textsuperscript{2}Meta FAIR
\\
 \small{
   \textbf{Correspondence:} \href{mailto:maggiehuang@meta.com}{maggiehuang@meta.com}, \href{mailto:lunadong@meta.com}{lunadong@meta.com}
 }
}

\begin{document}
\maketitle
\begin{abstract}
Production Retrieval-Augmented Generation (RAG) pays retrieval latency and serving cost on every query -- even when the model already knows the answer. We present {\ConfRAG}, a confidence-guided strategy that invokes RAG only when a large language model (LLM) is unlikely to be correct, cutting external retrievals by 5-39\% and tail latency by up to 800ms (P90) in a real RAG system while matching the answer quality of always-on RAG. {\ConfRAG} is enabled by {\ConfQA}, a lightweight fine-tuning strategy (one epoch over 3K knowledge-graph facts) that teaches an LLM to answer only when confident and otherwise say "I am unsure", cutting hallucination rates from 20-40\% to 5\% or below across multiple factuality benchmarks. Two choices make it work in practice: a dampening prompt that discourages overconfident hallucinations, and training on atomic facts that generalizes confidence calibration to unseen domains and question types. At serving time, LLM-QA and RAG run in parallel with early-stopping, so confident answers return immediately. 
\end{abstract}

\section{Introduction}
\label{intro}
Despite the remarkable capabilities that Large Language Models (LLMs) have demonstrated, they still \textit{hallucinate factual statements}~\citep{zhou2021detectinghallucinatedcontentconditional, dziri2022hallucinationorigin, rawte2023hallucinationemergence,mishra2024hallucinationdetection, kalai2025whyLMhallucinate}. 
It is broadly accepted that factual content shall not be \textit{fabricated} or \textit{generated}, but anchored in \textit{internally parameterized neural} knowledge or \textit{externally recorded symbolic} content (knowledge graphs, webpages, etc). Significant progress has been made on both fronts: knowledge internalization via pre-training~\citep{grattafiori2024llama3herdmodels, kim2025pretrainknowlentropy} and external knowledge use via Retrieval-Augmented Generation (RAG) (surveyed in ~\citep{zhou2024RAGtrustworthiness, gupta2024RAGsurvey, Ni2025RAGsurvey}).
However, a critical question remains: {\em when should LLMs rely on parameterized knowledge versus external sources?} 

Ideally, an LLM would consult external sources only when it detects uncertainty in its own knowledge, but reliable recognition of these knowledge limits remains unsolved. Token-level confidence methods require monitoring hidden-state signals from intermediate activations, often impractical in deployment~\citep{jiang2023activeretrievalaugmentedgeneration, su2024dragindynamicretrievalaugmented, li2025etc}. Prompting LLMs to explicitly express confidence is undermined by the well-known over-confidence~\cite{simpleqa2024, ni-etal-2024-llms, xiong2024can}. Recent fine-tuning approaches for uncertainty calibration help but remain imperfect, either only modestly reducing hallucinations or substantially suppressing answer correctness~\cite{cheng2024aiassistantsknowdont, yang2024alignmenthonesty, grattafiori2024llama3herdmodels, zhang2024rtuninginstructinglargelanguage, kapoor2025largelanguagemodelstaught, Li2025USTUning}.


This paper studies a fine-tuning approach that trains LLMs to recognize low-confidence factual knowledge and trigger RAG accordingly. Several key intuitions let us improve over prior work on eliminating hallucinations and avoiding unnecessary RAG invocations, yielding three contributions.

\begin{figure}[h]
    \centering
    \includegraphics[width=0.9\columnwidth]{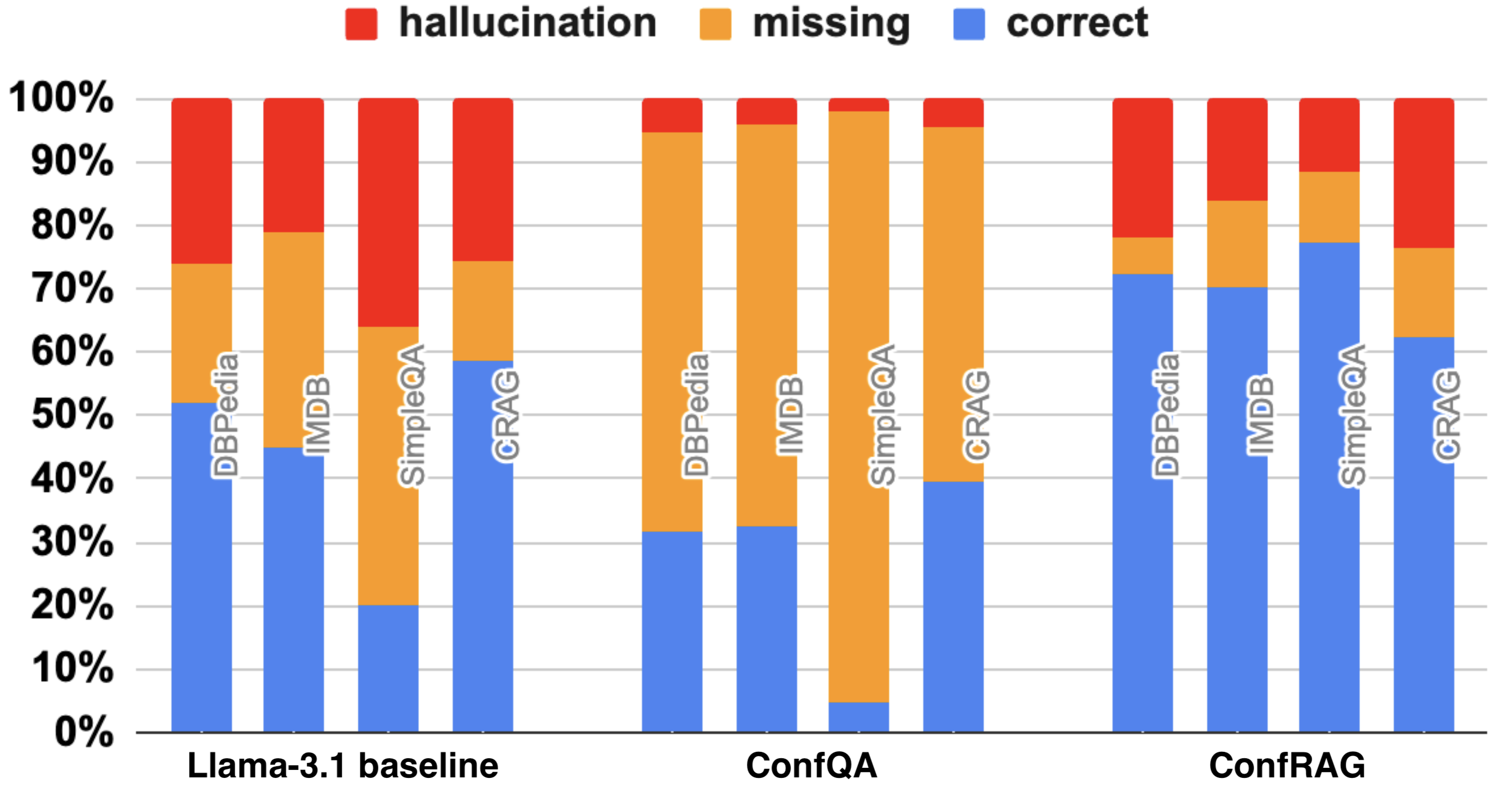}
    \caption{On factual QA benchmarks, fine-tuned model {\ConfQA} reduces hallucination to ~5\% or below; \ConfRAG\ increases accuracy by 57.2 points for SimpleQA and 3.6 points for CRAG while cutting latency (Table~\ref{tab:rag}).}
    \label{fig:overall_factuality}
\end{figure}

First, an effective fine-tuning method for confidence recognition, called {\sc ConfQA}, forms the core of our triggering strategy: it checks the LLM's inherent answer and teaches it to state ``I am unsure'' when the answer is incorrect. Two design choices distinguish it from existing work~\citep{zhang2024rtuninginstructinglargelanguage,cohen2024idontknowexplicit,cheng2024aiassistantsknowdont,yang2024alignmenthonesty,kapoor2025largelanguagemodelstaught}: (i) a {\em dampening} prompt---{\em ``Answer only if you are confident''}---that shapes the LLM's behavior, and (ii) training data of exclusively \textit{simple} entity attribute questions, whose atomic facts serve as building blocks for complicated statements, that let the confidence behavior generalize to broader domains. 


Second, a RAG triggering strategy called {\ConfRAG}, which casts triggering as detecting low-confidence answers: LLM-based QA and a RAG pipeline run in \textit{parallel}, with RAG halted early once the LLM gives a confident answer rather than ``I am unsure''. Unlike methods inspecting hidden-state signals~\citep{su2024dragindynamicretrievalaugmented, jiang2023activeretrievalaugmentedgeneration}, or prompt-instructed certainty responses~\citep{ni-etal-2024-llms}, {\ConfRAG} is lightweight, low latency, more accurate than baselines and broadly applicable.

Third, a comprehensive study across 8 benchmarks spanning short-form, long-form QA, and general knowledge QA. {\ConfRAG} reaches $\approx 95\%$ accuracy with a perfect RAG system, and with a real one matches always-on RAG accuracy while cutting P50 latency by over 600ms on benchmarks like CRAG~\citep{crag2024}. We also find two standalone uses of {\ConfQA} when RAG is unavailable (such as in Speech-in Speech-out systems~\citep{xie2024miniomnilanguagemodelshear,nguyen2022generativespokendialoguelanguage}): we recommend {\sc ConfQA} \textit{without} the dampener in inference to preserve accuracy with fewer hallucinations; we recommend {\ConfQA} \textit{with} the dampener, to drive hallucinations to under 5\% (see Figure~\ref{fig:overall_factuality}).
\section{Related Work}
\label{sec:related}
Our work relates to two bodies of work: LLM hallucination suppression and RAG triggering. 

\paragraph{LLM hallucination suppression:}LLMs are argued to be inherently prone to hallucinations due to the high-entropy next-token pre-training objective~\cite{kalai2025whyLMhallucinate}, prompting many mitigation efforts (surveyed in~\cite{tonmoy2024comprehensivesurveyhallucinationmitigation}). One direction incentivizes factually correct generation--leveraging internal parameters~\cite{li2023ITI, chuang2024DoLa}, reciting pre-training passages~\cite{Sun2023recitation}, or fine-tuning the model to select correct tokens~\cite{tian-etal-2023-just, xie2024fence, FLAME2024};
the other detects and suppresses likely-hallucinated content, e.g., verifying output tokens against internal or external corpus~\cite{lei2023CoNLI, Dhuliawala2023chainofverification, varshney2023stitch}.

{\ConfQA} belongs to the {\em confidence-based softening and abstaining} category, whose roots predate LLMs~\cite{kadavath2022language, mielke2022reducing, lin2022teachingmodelsexpressuncertainty}. Confidence has been derived via semantic entropy~\cite{kuhn2023semanticuncertaintylinguisticinvariances}, token-probability aggregation~\cite{duan2024SAR}, \textit{Local Intrinsic Dimension}~\cite{yin2024local-intrinsic-dimension}, self-reported confidence in verbal~\cite{zhou2023grayarea} or numerical~\cite{tian-etal-2023-just, xiong2024can} form, and consistency across generations~\cite{xiong2024can, yadkori2024mitigatingllmhallucinationsconformal} or models~\cite{zhang-etal-2023-sac3, feng2024donthallucinateabstainidentifying}. Given LLMs' well-known over-confidence, recent works fine-tune LLMs to abstain on low-confidence facts using three training signals: token probability~\cite{cohen2024idontknowexplicit, chen-etal-2025-teaching}, generation consistency~\cite{cheng2024aiassistantsknowdont, yang2024alignmenthonesty}, and answer against external references~\cite{grattafiori2024llama3herdmodels, zhang2024rtuninginstructinglargelanguage, kapoor2025largelanguagemodelstaught, Li2025USTUning}. 

\ConfQA\ also uses supervised fine-tuning (SFT) based on answer correctness, but our empirical study yields two surprising findings that motivate suppressing hallucinations without materially sacrificing accuracy. (i) a dampener prompt matters in both training and inference, and
(ii) training on atomic factual statements best teaches confidence and generalizes to unseen domains. Together these findings reduce hallucinations by an additional 5-11\% and improve factuality by up to 25\% over correctness-based SFT like~\cite{zhang2024rtuninginstructinglargelanguage}; we further find that adding a consistency constraint on top of correctness like~\cite{cheng2024aiassistantsknowdont} causes an unnecessary accuracy regression (Section~\ref{sec:5_finetune}).

\paragraph{RAG triggering:} 

RAG is widely studied and deployed to enhance LLMs' factuality, with numerous surveys
~\cite{Yu_2022,gao2024retrievalaugmentedgenerationlargelanguage,asai2024RAGsurvey, hu2024RAGRAU, zhao2024RAGsurvey, fan2024surveyragmeetingllms,huang2024surveyretrievalaugmentedtextgeneration, zhou2024RAGtrustworthiness, peng2024graphRAG, gupta2024RAGsurvey, Ni2025RAGsurvey}.
Modular RAG pipelines must optimize triggering (e.g., ETC~\cite{li2025etc}), retrieval (e.g., DeepRetrieval~\cite{jiang2025DeepRetrieval}), and noise-robust answer generation (e.g., InstructRAG~\cite{wei2025instructRAG}); \ConfRAG\ focuses on {\bf when} to trigger RAG.


RAG triggering shares its core intuitions with confidence-based abstention--withholding low-confidence facts in favor of external references--so related strategies overlap. \textit{Token-level} approaches need internal generation confidence, triggering RAG when next token has low certainty~\cite{he2021NNLM, jiang2023activeretrievalaugmentedgeneration} or high entropy~\cite{su2024dragindynamicretrievalaugmented, li2025etc}; {\em fact-level} approaches trigger RAG on low fact-level confidence~\cite{ni-etal-2024-llms, ren2025coling}. {\ConfRAG} is a fact-level approach, but differs in two aspects: it uses a fine-tuned model (\ConfQA) with low hallucination and only mild correction regression to best trade off quality and cost, and it parallelizes to hide the latency of model-based confidence testing. 

Two other triggering lines: (1) iterative methods that interleave retrieval and generation until the retrieved content suffices~\cite{peng2023checkfactstryagain, trivedi2023IRCoT, jeong2024AdaptiveRAG, zhu2025REAP}; (2) R1-style generation that emits a <search> token to trigger retrieval, then continues generation over the appended results~\cite{asai2023selfraglearningretrievegenerate, jin2025searchR1, li2025searchO1, chen2025research}. Neither targets the quality-latency trade-off of triggering. Our method is complementary and can improve their performance, either through training-data generation, or a curriculum-learning mechanism~\cite{sheng2018curriculumnet}.


\section{Methodology}
\label{sec:method}

We first describe \ConfQA, a fine-tuning approach for calibrating model confidence, and then show how it supports the triggering strategy.


\begin{figure}[t!]
    \centering
    \includegraphics[width=0.95\columnwidth]{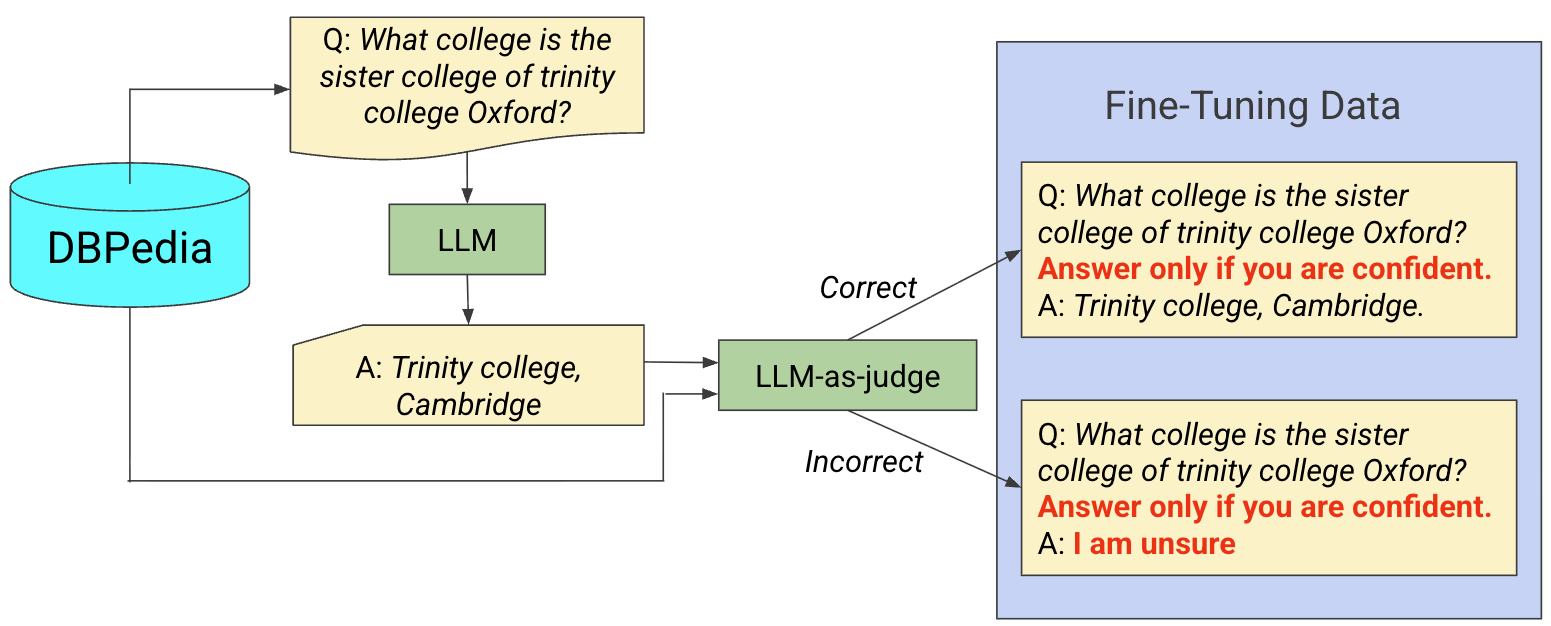}
    \caption{Training data generation for {\sc ConfQA} SFT.}
    \label{fig:training}
\end{figure}

\subsection{\ConfQA: When to Say Unsure?}
\label{subsec:model}
{\ConfQA} fine-tunes an LLM to answer only questions that it is confident about and to admit \textit{``I am unsure''} otherwise, guided by three intuitions: (1) calibrate the LLM's confidence by showing the ground truth; (2) add a {\em dampener prompt}, \textit{``Answer only if you are confident''}, to steer behavior explicitly; and (3) teach on {\em atomic facts} (entity attributes) so regularization targets factual statements.

{\bf Training data} are question–label pairs (see Figure~\ref{fig:training}). Questions ask for atomic facts and are generated from {\em DBPedia}, which spans diverse domains (Intuition \#3), using open-source script of~\cite{headtotail}, 
evenly sampled across entity popularity for representativeness. For labels, we prompt the Llama-3.1-70B model to answer each question (Prompt~\ref{prompt:short_form_gen_prompt} Appendix~\ref{sec:prompt}) and Llama-3.1-405B to judge the answer's consistency with the ground truth (Prompt~\ref{prompt:simple_qa_pair_grader} Appendix~\ref{sec:prompt}): correct answers are labelled with the ground truth, otherwise with \textit{``I am unsure about the answer''} (Intuition \#1).

The dampener prompt is supplied as the system prompt in both training and inference to suppress hallucinations (Intuition \#2). Appendix~\ref{appen:prompt_wording} shows rephrasing it has little impact.

\begin{figure}[t]
    \centering
    \includegraphics[width=0.95\columnwidth]{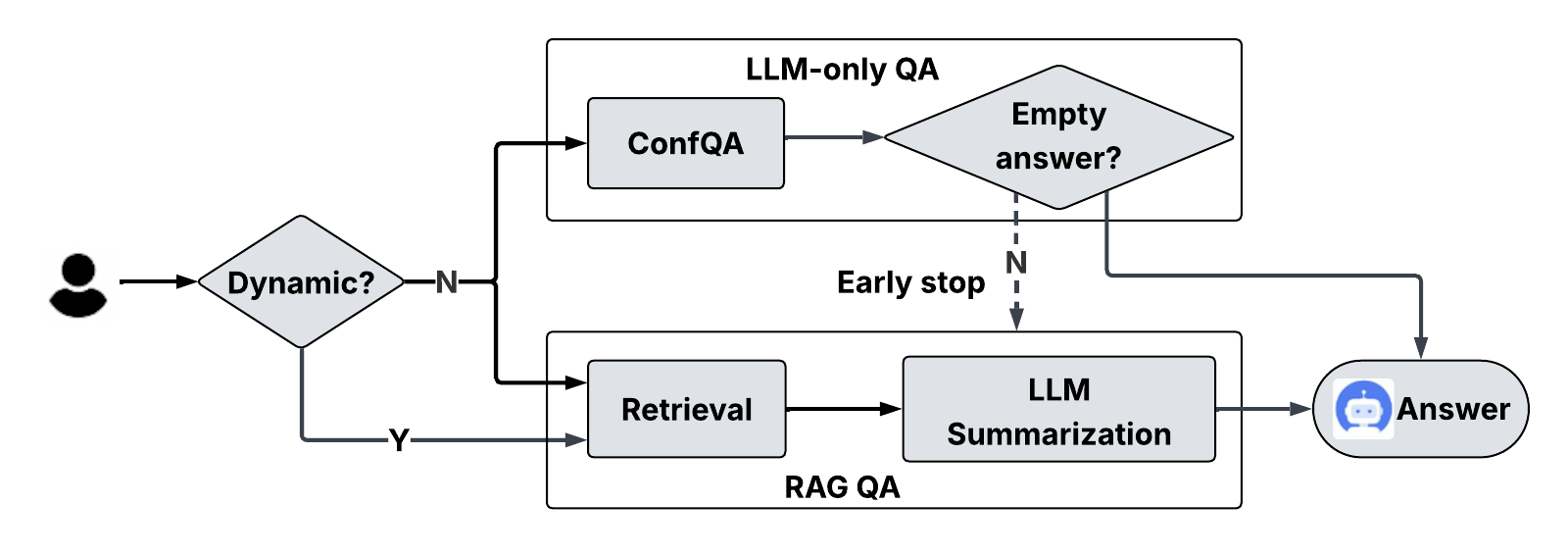}
    \caption{{\sc ConfRAG} invocation architecture.}
    \label{fig:parallel_sft_rag}
\end{figure}

\subsection{\ConfRAG: When to Trigger RAG?}
\label{subsec:trigger_rag}
Since {\ConfQA} answers only when confident, we invoke RAG only when it says "unsure". 

Figure~\ref{fig:parallel_sft_rag} depicts the {\ConfRAG} architecture, which extends {\em Factual Question Answering} to handle \textit{dynamic} questions, those needing up-to-date information, best answered by RAG. These are identified either by fine-tuning the same model or by a smaller dedicated model (Appendix~\ref{appen:dynamic}). For \textit{static} questions, {\ConfQA} and RAG run in parallel: if the LLM produces a valid answer, RAG is early-stopped and that answer is returned; otherwise, the system waits for and returns the RAG output.

\begin{table*}[t!]
\centering
\resizebox{1.7\columnwidth}{!}{%
\begin{tabular}{llllr}
\toprule
\bf Benchmark & \bf Category & \bf Question types & \bf \# Domain & \bf Size \\
\midrule

Head-to-Tail~\citep{headtotail} & short-form & simple questions (attribute of entities) & dbpedia, imdb & 1,200 \\
SimpleQA~\citep{simpleqa2024} & short-form & general fact-seeking questions & multiple domains & 4,326 \\
CRAG~\citep{crag2024} & short-form & simple questions, reasoning questions & 5 domains & 642 \\ \hline
LongFact~\citep{Wei24:longfact} & long-form & general questions & 38 domains & 250 \\
AlpacaFact~\citep{Dubois23:AlpacaFarm} & long-form & fact-seeking instruction-following & multiple domains & 241 \\
Biography~\citep{min2023factscorefinegrainedatomicevaluation} & long-form & biography questions & celebrity & 183 \\ \hline
MMLU 5-shot~\citep{hendrycks2021measuringmassivemultitasklanguage} & general knowl. & multi-choice questions & 57 domains & 14,042 \\
MMLU pro~\citep{wang2024mmlu-pro} & general knowl. & multi-choice questions & multiple domains & 12,032 \\
\bottomrule
\end{tabular}
}
\caption{The overall statistics of evaluation datasets.}
\label{tab:data_stats}
\end{table*}
\section{Experiment Setup}
\label{sec:data}

\subsection{Benchmarks and Metrics}
\label{subsec:benchmarks}
\paragraph{Data sets:}
We experimented with three {\em short-form factuality benchmarks}, where answers are mostly short. Questions range from {\em simple} questions about an entity's attribute to {\em complex} ones requiring comparison, aggregation, reasoning, and post-processing. Table~\ref{tab:data_stats} summarizes the benchmarks and Appendix~\ref{appen:data} gives details. Appendix~\ref{appen:confidence} shows that on all three, well-known LLMs are consistently over-confident in their answers, while answer consistency correlates better with accuracy--matching prior observations~\cite{mielke2022reducing, zhou2023grayarea, xiong2024can}.

\paragraph{Metrics and evaluation:}
For {\em QA quality}, following CRAG~\citep{crag2024}, we compute the percentage of {\em correct}, {\em incorrect} (hallucinations), and {\em missing} (not attempted) answers, and take {\em Factuality} = correct\% - incorrect\% as our main metric. Ranging from -1 to 1, Factuality penalizes hallucinations more than missing answers, letting us quantify hallucination mitigation as advocated in~\cite{kalai2025whyLMhallucinate}; Appendix~\ref{appen:data} further shows it penalizes hallucinations better than the usual \textit{F-measure}. We judge answer correctness via prompt-based LLM-as-a-judge, which was shown to achieve ~99\% accuracy in~\cite{crag2024}.

For {\em triggering}, we report precision and recall against an \textit{oracle} that triggers RAG whenever the LLM's answer from internal knowledge is not mostly correct (correct <2 out of 3 times).
{\textit{Precision}} is the fraction of triggered queries that the oracle would also trigger (i.e., how many of our RAG calls were necessary); {\textit{Recall}} is the fraction of oracle-trigger queries that we actually triggered (i.e., how many necessary calls we caught).
\textit{F-measure} computes their harmonic mean: $F_{msr} = {2 \cdot prec \cdot rec \over prec + rec}$.



\subsection{Models and Implementations}
\label{subsec:implementation}

\paragraph{Implementations:}
We fine-tuned Llama-3.1-70B~\cite{grattafiori2024llama3herdmodels}. Guided by a scaling-law study (see Appendix~\ref{appen:scaling-law}), we ran one epoch on 3K training samples, with learning rate 1e-6, batch size 1, and gradient accumulation step 1. Experiments used Nvidia H100 96GB HBM2e GPUs: 32 GPUs for fine-tuning, 8 for inference. We observed similar results on Qwen2.5-7B-Instruct~\cite{qwen2025qwen25technicalreport} and Gemma-3-4B-IT~\cite{gemmateam2025gemma3technicalreport}, see Appendix~\ref{appen:other_model}.

\paragraph{Alternative Solutions:}
We compared against two baselines: the oracle from Section~\ref{subsec:benchmarks}, and Llama-3.1-70B with an inference-time dampener--plus three state-of-the-art solutions: 

\smallskip
\noindent
{\em Prompt-based}~\citep{ni-etal-2024-llms, ren2025coling} triggers RAG when the LLM signals low confidence. We implemented both the vanilla version, and an enhanced one (Prompt+) urging prudence and improving correctness (punish+explain in~\citep{ni-etal-2024-llms}; Prompt~\ref{prompt:vanilla} and~\ref{prompt:punish_explain} in Appendix~\ref{sec:prompt}).

\smallskip
\noindent
{\em R-Tuning}~\citep{zhang2024rtuninginstructinglargelanguage} builds training data by appending \textit{``Are you sure you accurately answered the question based on your internal knowledge?''} in the question, and padding the ground truth with \textit{``I am sure''} or \textit{``I am unsure''} by correctness. We trigger RAG on an \textit{``I am unsure''} suffix. We trained on both MMLU~\citep{hendrycks2021measuringmassivemultitasklanguage} and DBPedia. 

\smallskip
\noindent
{\em IDK}~\citep{cheng2024aiassistantsknowdont} requires answer consistency (here at least four out of five times) as well as correctness when labeling ground truth. We fine-tuned on DBPedia for a more direct comparison.

\begin{table*}[t!]
\centering
\resizebox{1.7\columnwidth}{!}{%

\begin{tabular}{l|rrrr|rrrrr|}

\toprule
\bf Model & \bf Corr & \bf Miss & \bf Incor & \bf Fac. & \bf Ceiling Corr & \bf Ceiling Fac. & \bf Tri. Prec & \bf Tri. Rec   & \bf F$_{msr}$   \\
\midrule
 & \multicolumn{9}{c|}{\textbf{DBpedia} (in-domain)} \\

\cmidrule(rl){2-5} \cmidrule(rl){6-10} 

Llama-3.1 Oracle & \textit{47.2} & \textbf{21.5} & 31.3 & 15.9 & 68.7 & 37.4 & - & - & - \\
Llama-3.1 Dampen & 47.0 & \textit{26.8} & 26.2& 20.8 & 73.8 & 47.6 & \textit{94.4} & 47.9 & 63.3\\

\cmidrule(rl){2-5} \cmidrule(rl){6-10} 
Prompt & 28.7 & 56.3 & 15.0 & 13.7 & 85.0 & 70.0 & 68.0 & 72.6 & 70.2 \\
Prompt+ & 12.8 & 84.0 & \textit{3.2} & 9.6 & \textit{96.8} & \textit{93.6} & 60.1 & \textit{95.6} & 73.8 \\

\cmidrule(rl){2-5} \cmidrule(rl){6-10} 
R-tuning (MMLU) & 24.3 & 67.3 & 8.3 & 16.0 & 91.6 & 83.3 & 66.6 & 84.9 & 74.6 \\
R-tuning (DBPedia) & 24.5 & 67.8 & 7.7 & 16.8 & 92.3 & 84.6 & 67.3 & 86.4 & 75.7 \\
IDK (DBPedia) &  17.0 & 81.5 & \textbf{1.5} & 15.5 & \textbf{98.5} & \textbf{97.0} & 63.4 & \textbf{97.8} & 76.9 \\

\cmidrule(rl){2-5} \cmidrule(rl){6-10}  
Llama-3.1 20-consis & - & - & - & - & - & - & 74.6 & 90.9 & \textbf{81.9} \\
\cmidrule(rl){2-5} \cmidrule(rl){6-10} 
{\ConfQA\ (no dampener)} & \textbf{49.0} & 33.5 & 17.5 & \textbf{31.5} & 82.5 & 65.0 & \textbf{95.5} & 60.6 & 74.1\\
{\ConfQA} & 31.5 & 63.3 & 5.2 & \textit{26.3} & 94.8 & 89.6 & 73.9 & 88.6 & \textit{80.6} \\
\toprule
& \multicolumn{9}{c|}{\textbf{IMDb} (out-of-domain)} \\

\cmidrule(rl){2-5} \cmidrule(rl){6-10} 

Llama-3.1 Oracle & \textit{41.0} & \textbf{33.3} & 25.7 & 15.3 & 74.3 & 48.6 & - & - & - \\
Llama-3.1 Dampen & 40.7& \textit{36.2} & 23.2& 17.5 & 76.9 & 53.7 & \textbf{97.7} & 59.9 & 74.3 \\

\cmidrule(rl){2-5} \cmidrule(rl){6-10} 
Prompt & 31.8 & 54.3 & 13.8 & 18.0 & 86.1 & 72.3 & 85.0 & 78.2 & 81.5 \\
Prompt+ & 19.8 & 78.8 & \textit{1.3} & 18.5 & \textit{98.6} & \textit{97.3} & 73.4 & \textit{98.0} & 83.9 \\

\cmidrule(rl){2-5} \cmidrule(rl){6-10} 
R-tuning (MMLU) &28.2 & 60.5 & 11.3 & 16.9 & 88.7 & 77.4 & 80.2 & 82.2 & 81.2 \\
R-tuning (DBPedia) & 25.3 & 70.2 & 4.5 & 20.8 & 95.5 & 91.0 & 80.3 & 95.5 & 87.2 \\
IDK (DBPedia) &22.0 & 77.0 & \textbf{1.0} & 21.0 & \textbf{99.0} & \textbf{98.0} & 76.2 & \textbf{99.4} & 86.3 \\

\cmidrule(rl){2-5} \cmidrule(rl){6-10}  
Llama-3.1 20-consis & - & - & - & - & - & - & 87.8 & 95.2 & \textit{91.3} \\

\cmidrule(rl){2-5} \cmidrule(rl){6-10} 
{\ConfQA\ (no dampener)} & \textbf{43.0} & 42.0 & 16.0 & \textit{27.0} & 85.0 & 69.0 & \textit{96.4} & 68.1 & 79.8 \\
{\ConfQA} & 32.5 & 63.3 & 4.2 & \textbf{28.3} & 95.8 & 91.6 & 89.2 & 95.8 & \textbf{92.4} \\
\toprule

 & \multicolumn{9}{c} {\textbf{SimpleQA} (out-of-domain)} \\

\cmidrule(rl){2-5} \cmidrule(rl){6-10}  

Llama-3.1 Oracle &16.0 & \textit{42.7} & 41.3 & -25.3 & 58.7 & 17.4 & - & - & - \\
Llama-3.1 Dampen &16.8& 48.0 &35.2&-18.4 & 64.8 & 29.6 & \textbf{97.7} & 55.8 & 71.0 \\

\cmidrule(rl){2-5} \cmidrule(rl){6-10} 
Prompt & 0.9 & 97.6 & \textit{1.5} & -0.6 & \textit{98.5} & \textit{97.0} & 84.3 & 97.9 & 90.6 \\
Prompt+ & 1.5 & 96.8 & 1.7 & -0.2 & 98.3 & 96.6 & 85.1 & \textit{98.0} & 91.1 \\

\cmidrule(rl){2-5} \cmidrule(rl){6-10} 
R-tuning (MMLU) & \textbf{20.3} & \textbf{38.0} & 41.7 & -21.4 & 58.3 & 16.6 & 87.2 & 88.4 & 87.8 \\
R-tuning (DBPedia) &3.7 & 83.3 & 13.0 & -9.3 & 87.0 & 74.0 & 85.4 & 84.7 & 85.1 \\
IDK (DBPedia) & 0.6 & 99.1 & \textbf{0.2} & \textit{0.4} & \textbf{99.7} & \textbf{99.5} & 84.4 & \textbf{99.7} & 91.4 \\

\cmidrule(rl){2-5} \cmidrule(rl){6-10} 
Llama-3.1 20-consis & - & - & - & - & - & - & 88.9 & 95.2 & \textit{92.0} \\

\cmidrule(rl){2-5} \cmidrule(rl){6-10} 
{\ConfQA\ (no dampener)} & \textit{17.3} & 55.8 & 26.8 & -9.5 & 73.1 & 46.3 & \textit{97.3} & 64.7 & 77.7 \\
{\ConfQA} & 4.9 & 93.1 & 2.1 & \textbf{2.8} & 98.0 & 95.9 & 87.6 & 97.1 & \textbf{92.1} \\
\toprule

 & \multicolumn{9}{c} {\textbf{CRAG} (out-of-domain)}  \\

\cmidrule(rl){2-5} \cmidrule(rl){6-10} 

Llama-3.1 Oracle & \textbf{60.3} & \textbf{11.7} & 28.0 & 32.3 & 72.0 & 44.0 & - & - & - \\
Llama-3.1 Dampen & 57.5&22.3&20.2&\textbf{37.2} & 79.8 & 59.6 & \textbf{83.9} & 47.1 & 60.3 \\

\cmidrule(rl){2-5} \cmidrule(rl){6-10} 
Prompt & 23.7 & 69.3 & 7.0 & 16.7 & 93.0 & 86.0 & 48.1 & 83.9 & 61.1 \\
Prompt+ & 26.5 & 71.2 & \textit{2.3} & 24.2 & \textit{97.7} & \textit{95.4} & 53.2 & \textit{95.3} & 68.3 \\

\cmidrule(rl){2-5} \cmidrule(rl){6-10} 
R-tuning (MMLU) &  \textit{57.8} & \textit{17.1} & 25.1 & 32.7 & 74.9 & 49.8 & 51.2 & 72.9 & 60.2 \\
R-tuning (DBPedia) & 31.6 & 55.0 & 13.4 & 18.2 & 86.6 & 73.2 & 50.7 & 70.2 & 58.9 \\
IDK (DBPedia) & 20.7 & 78.2 & \textbf{1.1} & 19.6 & \textbf{98.9} & \textbf{97.8} & 49.8 & \textbf{98.0} & 66.1 \\

\cmidrule(rl){2-5} \cmidrule(rl){6-10} 
Llama-3.1 20-consis & - & - & - & - & - & - & 65.9 & 80.4 & \textit{72.4} \\

\cmidrule(rl){2-5} \cmidrule(rl){6-10}  
{\ConfQA\ (no dampener)} & 57.0 & 19.6 & 23.4 & 33.6 & 76.6 & 53.2 & \textit{83.3} & 41.2 & 55.1 \\
{\ConfQA} & 39.4 & 56.2 & 4.4 & \textit{35.0} & 95.6 & 91.2 & 63.7 & 90.2 & \textbf{74.7} \\
\bottomrule

\end{tabular}
}
\caption{Overall factuality and RAG triggering F$_{msr}$ (all numbers in percentage \%; bold indicates the best performance and italic the second best). {\ConfQA} improves factuality and reduces incorrectness to ~5\% and below, achieving highest triggering F$_{msr}$ on 3 benchmarks and close to highest triggering F$_{msr}$ on DBPedia (Llama-3.1 20-consis). Note: Llama-3.1 20-consis calls LLM 21 times, with 20 response generations and 1 consistency judge. Not practical in real deployment.} 
\label{tab:overall}
\end{table*}

\section{Experimental Results}
\label{sec:5_finetune}
In this section, we study the effectiveness of \ConfQA\ and \ConfRAG. %

\subsection{Effectiveness of {\ConfQA}}
\label{subsec:short-form}
The left part of Table~\ref{tab:overall} reports {\ConfQA}'s answer quality. First, baseline methods are hallucination prone. The dampener cut incorrect answers by only 2.5-7.8 points across benchmarks.

\smallskip
Second, prompt-based methods suppress hallucinations, but hurt correctness, sometimes lowering factuality. The vanilla version still incurs high incorrectness (DBPedia, IMDb); while the stricter enhanced version suppresses hallucinations effectively, but sharply reduced correctness.

\smallskip
Third, R-tuning methods suppress hallucinations but hurt correctness. R-tuning on DBPedia suppresses hallucinations more effectively than on MMLU, but hurt correctness more. R-tuning on MMLU has high incorrectness rate--sometimes above baseline--for two reasons: 1) feeding ground truths as extra knowledge can encourage hallucinating what was not learned in pre-training; 2) MMLU mixes facts with reasoning, introducing ambiguity. This supports our hypothesis that atomic facts yield better training examples (Intuition \#3). 

\smallskip
Fourth, IDK achieves the lowest incorrectness rate (below 1.5\% on all benchmarks) but drops correctness significantly, reducing overall factuality--showing that additionally requiring the consistency signal can be unnecessarily strict.

\smallskip
Fifth, fine-tuned {\ConfQA} improves factuality by up to 25\%, with incorrectness dropping to ~5\% or below on all benchmarks. Correctness drops too, but far less than hallucinations drop, yielding much higher factuality. Although trained only on DBPedia, {\ConfQA} generalizes to other benchmarks (Intuition \#3). 


\begin{table*}[h!]
\centering
\resizebox{1.7\columnwidth}{!}{%
\begin{tabular}{l|rrrrrrr|rrrrrrr}

\toprule
\multirow{2}{*}{\bf Model} & \multicolumn{7}{c|}{\textbf{SimpleQA}} & \multicolumn{7}{c} {\textbf{CRAG}}\\

\cmidrule(rl){2-8} \cmidrule(rl){9-15} 

& \bf Upper & \bf Corr & \bf Miss & \bf Incor & \bf Fac & \bf L-P50 & \bf L-P90 & \bf Upper & \bf Corr & \bf Miss & \bf Incor & \bf Fac & \bf L-P50 & \bf L-P90\\
\midrule

LLM-only & 20.0 & 20.0 & 44.1 & 35.9 & -15.8 & 480 & 896 & 58.7 & 58.7 & 15.6 & 25.7 & 33.0 & 480 & 896 \\
RAG-everywhere & \textbf{100.0} & \textbf{78.1} & 11.5 & \textbf{10.5} & \textbf{67.6} & 1,900 & 2,780 & \textbf{100.0} & 61.1 & 15.1 & 23.8 & 37.3 & 1,900 & 2,780 \\
{\ConfRAG} & 95.1 & 77.2 & \textbf{11.4} & 11.5 & 65.7 & 1,802 & 2,650 & 95.6 & \textbf{62.3} & \textbf{14.2} & \textbf{23.5} & \textbf{38.8} & 1,278 & 1,955\\
\bottomrule
\end{tabular}
}
\caption{{\ConfRAG} achieves similar quality to RAG-everywhere, while cutting latency. {\bf Upper}: upper bound of percentage of correct. {\bf L-P50}: P50 latency and {\bf L-P90}: P90 latency.}
\label{tab:rag}
\end{table*}

\begin{table*}[h!]
\centering
\resizebox{1.7\columnwidth}{!}{%
\begin{tabular}{l|rrrr|rrrr|rrrr|rrrr|}

\toprule
\multirow{2}{*}{\bf Model} & \multicolumn{4}{c|}{\textbf{DBPedia} (in-domain)} & \multicolumn{4}{c|} {\textbf{IMDb} (out-of-domain)} & \multicolumn{4}{c|} {\textbf{SimpleQA} (out-of-domain)}& \multicolumn{4}{c|} {\textbf{CRAG} (out-of-domain)}  \\

\cmidrule(rl){2-5} \cmidrule(rl){6-9} \cmidrule(rl){10-13} \cmidrule(rl){14-17} 

& \bf Corr & \bf Miss & \bf Incor & \bf Fac & \bf Corr & \bf Miss & \bf Incor & \bf Fac & \bf Corr & \bf Miss & \bf Incor & \bf Fac & \bf Corr & \bf Miss & \bf Incor & \bf Fac \\
\midrule

Llama-3.1 Dampen& 47.0 & 26.8& 26.2& 20.8& 40.7& 36.2& 23.2& 17.5
&16.8&48.0&35.2&-18.4& \textbf{57.5} &22.3&20.2&\textbf{37.2} \\
\midrule

{\ConfQA} & 31.5 & 63.3 & 5.2 & \textbf{26.3} & 32.5 & 63.3 & 4.2 & \textbf{28.3}
&4.9&93.1& 2.1 & \textbf{2.8} &39.4 &56.2 & 4.4 &35.0 \\

* No-dampener in Train & 36.0& 50.2 & 13.8& 22.2& 
34.2& 50.7& 15.2& 19.0& 
5.8& 87.5& 6.7& -0.9& 
46.0& 44.2& 9.8& 36.2\\

* MMLU-as-source & 8.2 & 89.8 & \textbf{2.0} &6.2 & 16.7 & 82.0 & \textbf{1.3} & 15.4 &
0.6 & 98.8 & \textbf{0.5} & 0.1 & 7.0 & 92.7 &\textbf{0.3} & 6.7 \\

* GT-as-label & \textbf{48.0} & \textbf{2.8} & 49.2 & -1.2 & \textbf{41.2} & \textbf{4.3} &  54.5 & -13.3 & \textbf{17.8} & \textbf{13.7} & 68.4 & -50.6 & 53.7 & \textbf{14.2} &  32.1 & 21.6 \\

\bottomrule
\end{tabular}
}
\caption{Ablation study for {\ConfQA} when applying dampener in inference for all models, showing effectiveness of our fine-tuned model. All numbers are in percentage~(\%) and full results in Table~\ref{tab:ablation-full}.}
\label{tab:ablation}
\end{table*}

\begin{table*}[h!]
\centering
\scriptsize
\centering
\resizebox{1.65\columnwidth}{!}{%
\begin{tabular}{l|rrrr|rrrr|rrrr}
\toprule
\multirow{2}{*}{\bf Model} & 
\multicolumn{4}{c|}{\textbf{Long Fact}} & 
\multicolumn{4}{c|}{\textbf{Alpaca Fact}} & 
\multicolumn{4}{c}{\textbf{Biography}} \\

\cmidrule(rl){2-5} \cmidrule(rl){6-9} \cmidrule(rl){10-13}

& \bf Prec & \bf Rec & \bf F1 & \bf Miss & \bf Prec & \bf Rec & \bf F1 & \bf Miss & \bf Prec & \bf Rec & \bf F1 & \bf Miss\\
\midrule

Llama3.1 & 64.5 & 65.4 & 64.3 & \textbf{0} & \textbf{62.3} & 71.0 & \textbf{63.8} & \textbf{0} & 35.4 & 40.3 & 37.1 & \textbf{0}  \\
{\ConfQA} & \textbf{67.0} & \textbf{67.7} & \textbf{66.7} & 0.8 & 62.2 & \textbf{71.1} & \textbf{63.8} & 0.4 & \textbf{42.0} & \textbf{46.5} & \textbf{42.6} & 12.6 \\
\bottomrule
\end{tabular}
}
\caption{{\ConfQA} matches or exceeds baseline for long-form generation.}
\label{tab:longformgen}
\end{table*}

\subsection{\ConfRAG\: Triggering and Serving}
\label{subsec:6_trigger}
The right part of Table~\ref{tab:overall} compares triggering metrics across solutions. As a strong but impractical reference point, we include Llama-3.1 20-consis, which generates an answer 20 times per question and then calls Llama-3.1 once more to judge whether the 20 responses are mutually consistent; it triggers RAG unless all 20 responses agree and none is a refusal. Because this consistency signal is not an answer-quality decision, Llama-3.1 20-consis has no entries in the left (answer-quality) part of Table 2. It is also far costlier than {\ConfRAG}--$\approx$20× the generation tokens and at least twice the latency--which makes it unsuitable for deployment. 

\smallskip
{\ConfRAG} lifts potential correctness to $\approx$ 95\%, while cutting unnecessary external retrievals by 5-39\%. It attains the highest triggering F-measure on every benchmark except DBPedia, where Llama-3.1 20-consis is highest; even there, {\ConfRAG} is within a small margin--so {\ConfRAG} matches or exceeds Llama-3.1 20-consis on triggering quality at a fraction of its token and latency cost. The remaining methods fall short on one of two axes: Prompt and R-tuning have low ceiling correctness, while Prompt+ and IDK reach a higher ceiling but at much lower triggering precision, incurring unnecessary retrieval cost.

\paragraph{Deployment architecture:} at serving time (Figure~\ref{fig:parallel_sft_rag}), LLM-QA and the RAG pipeline run in parallel; once {\ConfQA} returns a confident answer, retrieval is early-stopped and the answer returned, otherwise the system waits for RAG. Because {\ConfQA} abstains only on genuinely uncertain queries, {\ConfRAG} finishes retrievals on a fraction of traffic, cutting unnecessary external retrievals by 5-39\% (Table~\ref{tab:overall}) and end-to-end latency by 600ms (P50) and 800ms (P90) on CRAG (Table~\ref{tab:rag}). SimpleQA requires triggering RAG for most questions to reach high quality, so its latency gain is smaller. For a service handling $N$ queries/day at retrieval unit cost $c$, this avoids $\approx$(0.05 to 0.39)$\cdot N \cdot c$ in retrieval spend plus the corresponding generator GPU-seconds and tokens.

\smallskip
Table~\ref{tab:rag} shows results of production implementation of {\ConfRAG} and RAG-everywhere for SimpleQA and CRAG benchmarks. {\ConfRAG} shows slight under-performance on SimpleQA factuality and Correctness and slight improvement on CRAG compared with always-on RAG. These numbers came from production implementation of a RAG system, so accounting for small variations, {\ConfRAG} is more or less on-par with always-on RAG for quality. 
On our production-representative pipeline (Bing + Knowledge Graph APIs feeding Llama-3.1-70B), retrieval and summarization are the dominant per-query cost.

\paragraph{RAG cost saving:} our production RAG pipeline runs four sequential stages: a search API call (500ms), crawling 10 pages (1300ms), content extraction and post-processing from HTML (400ms), and an LLM call over the extracted content to generate an answer (3000ms) -- 5200ms end to end. When {\ConfRAG} decides to answer directly, the search API is still triggered, but the pipeline halts during crawling, before post-processing and summarization. Because search runs in parallel with direct QA, its 500ms is fully overlapped and off the critical path; the three downstream stages are eliminated along with their cost: 1). Crawling (10 pages) -- interrupted mid-process, avoiding most of the 1300ms latency and the associated bandwidth/crawl cost. 2). Extraction, chunking, and re-ranking -- the full 400ms of post-processing and any paid API cost therein are saved.
3). LLM summarization -- with no page content produced, both the summarization call and the tokens it would consume are avoided.

{\bf The token savings are substantial}: summarizing 10 webpages to answer a question consumes roughly 15,000-50,000+ tokens, making the retrieval path 30-100× more expensive per query than direct QA.

\paragraph{Fallback when RAG is unavailable:} in offline or low-latency settings (on-device, Speech-in Speech-out), run {\ConfQA} without the dampener to preserve correctness while suppressing hallucinations; add the dampener to drive hallucinations under 5\% when maximum reliability is required.



\begin{table}
\centering
\resizebox{0.7\columnwidth}{!}{
        \begin{tabular}{l|r|r} \hline
        \toprule
          {\bf Model} & {\textbf{MMLU (5-shot)}} &   {\textbf{MMLU Pro}} \\
          \midrule				
          Llama3.1 & 82.7 & \textbf{66.3}  \\
          {\ConfQA} & \textbf{82.8} & 65.4   \\
          \bottomrule
        \end{tabular}
        }
        \captionof{table}{{\ConfQA} does not regress on MMLU.}
        \label{tab:mmlu_gen}
\end{table}

\subsection{Deep dive on {\ConfQA}}
\label{subsec:confqa}
Since the supervised model \ConfQA\ is critical to deciding when to trigger RAG, we now examine its design choices and generalizability. 

\paragraph{Ablation study:}
We compare {\ConfQA} with several alternatives in Table~\ref{tab:ablation} (full results in Table~\ref{tab:ablation-full} Appendix~\ref{appen:ablation}). First,
dropping the dampener from training raises correctness and lowers the missing rate, but increased hallucinations and thus lower factuality--confirming the dampener's importance in training (Intuition \#2). 
Second, generating training data from MMLU instead of DBPedia yields low hallucination ($<2\%$) but far lower correctness (Intuition \#3). 
Third, using the ground truth directly as the label ({\em GT-as-label}) gives the highest correctness and lowest missing rate, but hallucinates much more (up to 68.4\%), consistent with observations in previous work~\citep{FLAME2024,gekhman2024doesfinetuningllmsnew}.

\paragraph{Generalizability to long-form benchmarks:}
We test {\em long-form factuality benchmarks}, where answers contain multiple factual statements (see Table~\ref{tab:data_stats} and Appendix~\ref{appen:data}). We evaluate with {\em VeriScore}~\citep{song2024veriscoreevaluatingfactualityverifiable}, which computes precision, recall, and F1, using their fine-tuned models for claim extraction and verification; the minimum facts for perfect recall is set to the median extracted claims per dataset.
We omit the dampener prompt to avoid regressing recall.
Table~\ref{tab:longformgen} shows {\ConfQA} matches or exceeds baseline precision and recall, except on 13\% long-tail biography questions, which it declines to answer due to low confidence.

\paragraph{No regression on general benchmarks:}
Finally, we evaluated \ConfQA\ on MMLU's multi-choice questions on general knowledge and reasoning~\citep{hendrycks2021measuringmassivemultitasklanguage, wang2024mmlu-pro}, again without the dampener prompt. Table~\ref{tab:mmlu_gen} shows scores mostly similar to baseline. 

\section{Conclusion}


We presented {\ConfRAG}, a confidence-guided RAG architecture that triggers RAG only when an LLM is unlikely to be correct. Built on \ConfQA\ -- a lightweight fine-tune that teaches a model to answer when confident and otherwise say "I am unsure" -- it cuts hallucinations to 5\% or below and matches RAG-everywhere accuracy (slight under-performance on SimpleQA and slight improvement on CRAG) while reducing retrievals by 5-39\%.


{\ConfRAG} is practical to deploy and maintain: {\ConfQA} retrains in one epoch over 3K auto-generated knowledge-graph facts and transfers across model families (Appendix~\ref{appen:other_model}), requiring no access to model internals or per-domain labeling. As base models and retrievers improve, integrating internal and external knowledge this way should further improve both factuality and latency.

\newpage
\section{Limitations}
\label{sec:limitations}
Our experiments focus on SFT, and we leave DPO-based fine-tuning for future work. 
We empirically compared DBPedia and MMLU, where the former contains only simple factual questions, and the latter contains questions ranging from factual to reasoning. A comprehensive study regarding the effect of sources with different coverage in this spectrum can further improve understanding. 
SFT requires access to LLM parameters, thus restricting the application of the proposed framework to open-sourced models. Lastly, our cost/latency results come from a replay of the benchmark on production pipeline rather than a live A/B test, and savings depend on traffic mix and retrieval pricing.
\bibliography{custom}
\clearpage

\appendix

\section{Benchmark Setup Details}
\label{appen:data}
To evaluate the performance of the fine-tuning, We consider 8 benchmarks in the main content as described in Section~\ref{subsec:benchmarks} Table~\ref{tab:data_stats}, with details described here.
\begin{itemize}
\item \textbf{{\sc\bf Head-to-Tail} (\textit{short-form} with simple questions)}~\citep{headtotail}: We leveraged the data scraping scripts from the {\bf Head-to-Tail} benchmark, and generated simple questions and their answers according to content from DBPedia~\footnote{\url{dbpedia.org}} (general knowledge graph) and IMDb~\footnote{\url{imdb.com}} (data in the {\em Movie} domain). From each dataset we randomly sampled 200 entities for \textit{head} entities, 200 for \textit{torso} entities, and 200 for \textit{tail} entities. Here we follow the definition in~\cite{headtotail} for head, torso and tail: we rank all entities by their traffic; head entities are top-popular entities that together account for 1/3 of traffic, tail entities are unpopular entities that together account for 1/3 of traffic, and torso entities are the remaining medium-popular entities. Together, we have 1200 question-answer pairs, 600 from each source.

\item \textbf{{\sc\bf SimpleQA} (\textit{short-form} with simple questions)}~\citep{simpleqa2024}: {\bf SimpleQA} is a benchmark released by \textit{OpenAI} to measure LLM factuality. It contains 4326 manually crafted short, fact-seeking questions, covering diverse topics such as science, technology, history, and entertainment. 

\item \textbf{{\sc\bf CRAG} (\textit{short-form} with simple and complex questions)}~\citep{crag2024}: CRAG is a benchmark to test RAG capabilities. It contains 4,409 training and 1,335 evaluation questions covering five domains (general, finance, sports, music, movie), entities of different popularities (head, torso, tail), facts of different dynamisms (static, slow-changing, fast-changing, real-time), and eight question types (simple, condition, set, comparison, aggregation, multi-hop, post-processing, false premise). We selected the 642 {\em static} questions from the evaluation data set, with 97 questions for head entities, 99 for torso, 90 for tail entities and 356 for facts from the web (mostly popular); we excluded false-premise and dynamic questions from the sampling as it presents different challenges.

\item \textbf{{\sc\bf LongFact} (\textit{long-form})}~\citep{Wei24:longfact}: Aiming to measure the factuality of long-form responses consisting of at least several paragraphs, LongFact has 2,280 factual questions covering 38 topics, generated by prompting GPT-4. Following \citet{Wei24:longfact}, we use the 250 prompts from the LongFact-Objects dataset in our experiments.

\item \textbf{{\sc\bf AlpacaFact} (\textit{long-form})}~\citep{FLAME2024}: 
Initially sourced from diverse interactions with real-world users, the 805 instructions in AlpacaFarm~\citep{Dubois23:AlpacaFarm} served as a benchmark for evaluating the ability of different LLMs to follow instructions. Following ~\citet{FLAME2024}, we used a subset of 241 fact-seeking instructions in this work.

\item \textbf{{\sc\bf Biography} (\textit{long-form})}~\citep{min2023factscorefinegrainedatomicevaluation}: To validate the effectiveness of FActScore, ~\citet{{min2023factscorefinegrainedatomicevaluation}} created a collection of prompts named Biography by applying the template ``\texttt{Tell me a bio of [Person Name]}'' to 183 notable individuals listed on Wikipedia. Given its extensive use in recent literature, we have included this prompt set for our experiments as well.

\item \textbf{{\sc\bf MMLU} (\textit{General knowledge})}: The MMLU \citep{hendrycks2021measuringmassivemultitasklanguage} dataset covers 57 subjects, including areas such as mathematics, history, law, and medicine. It contains two subsets: the MMLU 5-shot dataset contains 14,042 multi-choice questions to evaluate general knowledge and problem-solving tasks; the MMLU Pro~\citep{wang2024mmlu-pro} dataset contains 12,032 multi-choice questions to stress-test reasoning, disambiguation, and factual accuracy.
\end{itemize}

For short-form questions we consider factuality score as defined in Section~\ref{subsec:benchmarks} rather than F1 score, where F1-score is more lenient for incorrect answers (hallucinations), but factuality strongly prefers missing answers to hallucinations. For example, consider a model that answers 10\% questions correctly (correct\% = 10\%) and the rest of the questions incorrectly (incorrect\% = 90\%); the F1-score is 10\% (not punishing hallucinations much) while the factuality is -80\%. Now consider another model that answers 10\% questions correctly and admits "I am unsure about the answer" for the rest of the questions; the F1-score is 18.2\%, only slightly higher than 10\%, but the factuality is 10\%, significantly higher than -80\%. Finally, consider a set of binary questions. A model that does not know any answer will get factuality of 0\% no matter whether it admits so or makes random guesses ($0=50\%-50\%$); however, if we use F-measure, the former gets a F-measure of 0 and the latter gets a F-measure of 50\%, encouraging random guesses and thus hallucinations. 

For long-form responses we use the automatic evaluation metric, {\em VeriScore}~\cite{song2024veriscoreevaluatingfactualityverifiable}, for measuring the factuality (results in Section~\ref{subsec:confqa} Table~\ref{tab:longformgen}). Following FActScore~\citep{min2023factscorefinegrainedatomicevaluation} and SAFE~\citep{Wei24:longfact}, VeriScore extracts more sensible and verifiable claims from each sentence and uses Google search snippet instead of Wikipedia as the source of knowledge. This approach allows VeriScore to be applied to more diverse topics and requires fewer but more meaningful claims to be checked. We report the F1 score from VeriScore, which represents the harmonic mean of precision and recall. In line with \cite{song2024veriscoreevaluatingfactualityverifiable}, we set the minimum number of facts required for perfect recall based on the median number of extracted claims per dataset, using their fine-tuned models for claim extraction and verification.

\section{Analysis of LLM's Self-Reported Confidence}
\label{appen:confidence}
Using three benchmarks, we reproduce the observations in~\cite{simpleqa2024,xiong2024can}, that LLMs' self-reported confidence is systematically overestimated, making it a poor signal for deciding when to trigger RAG.

\begin{figure*}[t]
\centering
\includegraphics[width=0.9\textwidth]{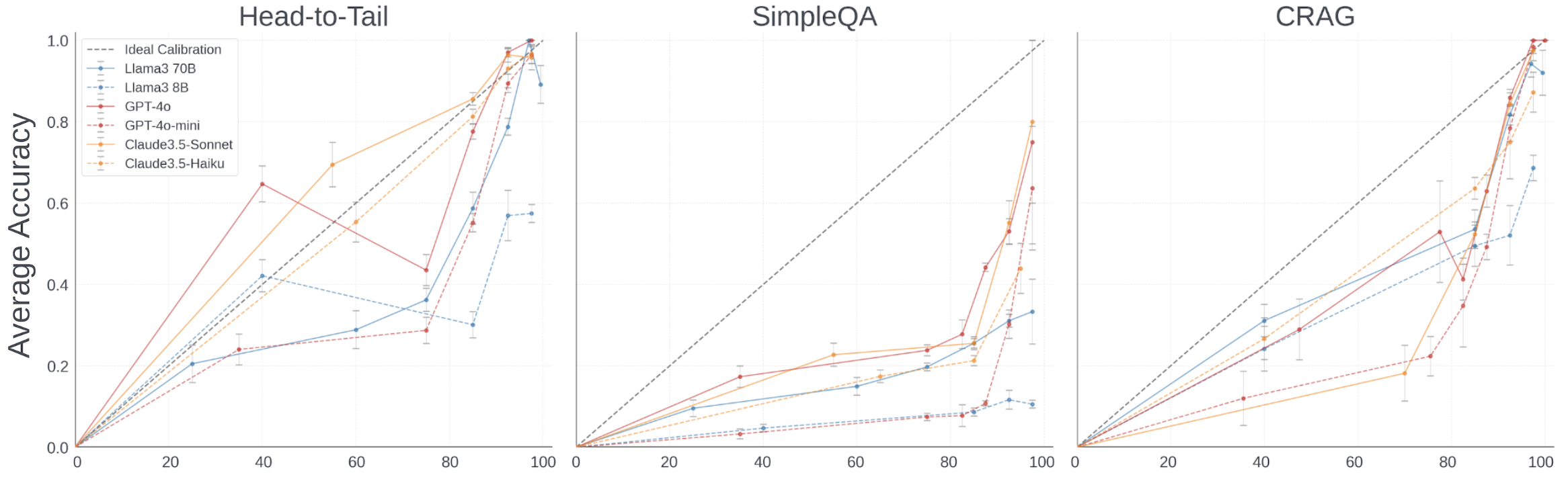}
\caption{LLMs' self-reported confidences are correlated with QA accuracy, but often over-confident.}
\label{fig:confidence-main}
\end{figure*}

We prompted the LLM to output a confidence score between 0 and 1 along with its answer (Prompt~\ref{prompt:self-confidence} in Appendix~\ref{sec:prompt}). 
After removing missing answers, we binned the reported confidences into equal-sized quantiles, and plotted average accuracy per bin to assess \textit{calibration}: whether a confidence of 0.8 corresponds to ~80\% accuracy? Figure~\ref{fig:confidence-main} shows the calibration, leading to three observations.

\begin{enumerate}
    \item The self-reported confidence is mostly positively correlated with QA accuracy, but LLMs tend to be over-confident (the correlation curves are below the ideal calibration dashed line). For example, when Llama-3.1-70B predicts a confidence of 80\% on CRAG, the real accuracy is only 33\%. 
    \item Notably, for the same model series, the smaller model is often more confident than the larger model (with an exception of Claude3.5 on CRAG), demonstrating the interesting correlation between ignorance and self-assurance.  
    \item Finally, the overconfidence is more pronounced when answering SimpleQA questions than on other benchmarks. A sample of 50 questions from SimpleQA shows that the questions are often nuanced for fairly popular entities 
    (e.g. \textit{``What was the first line after the salutation in the letter sent to Sardar Patel by Abhay Charan De?", ``In which month and year was Service Pack 3 for Windows Fundamentals for Legacy PCs released?''}), possibly causing LLMs to be over-confident. 
\end{enumerate}

\begin{figure*}[ht]
\centering
\includegraphics[width=\textwidth]{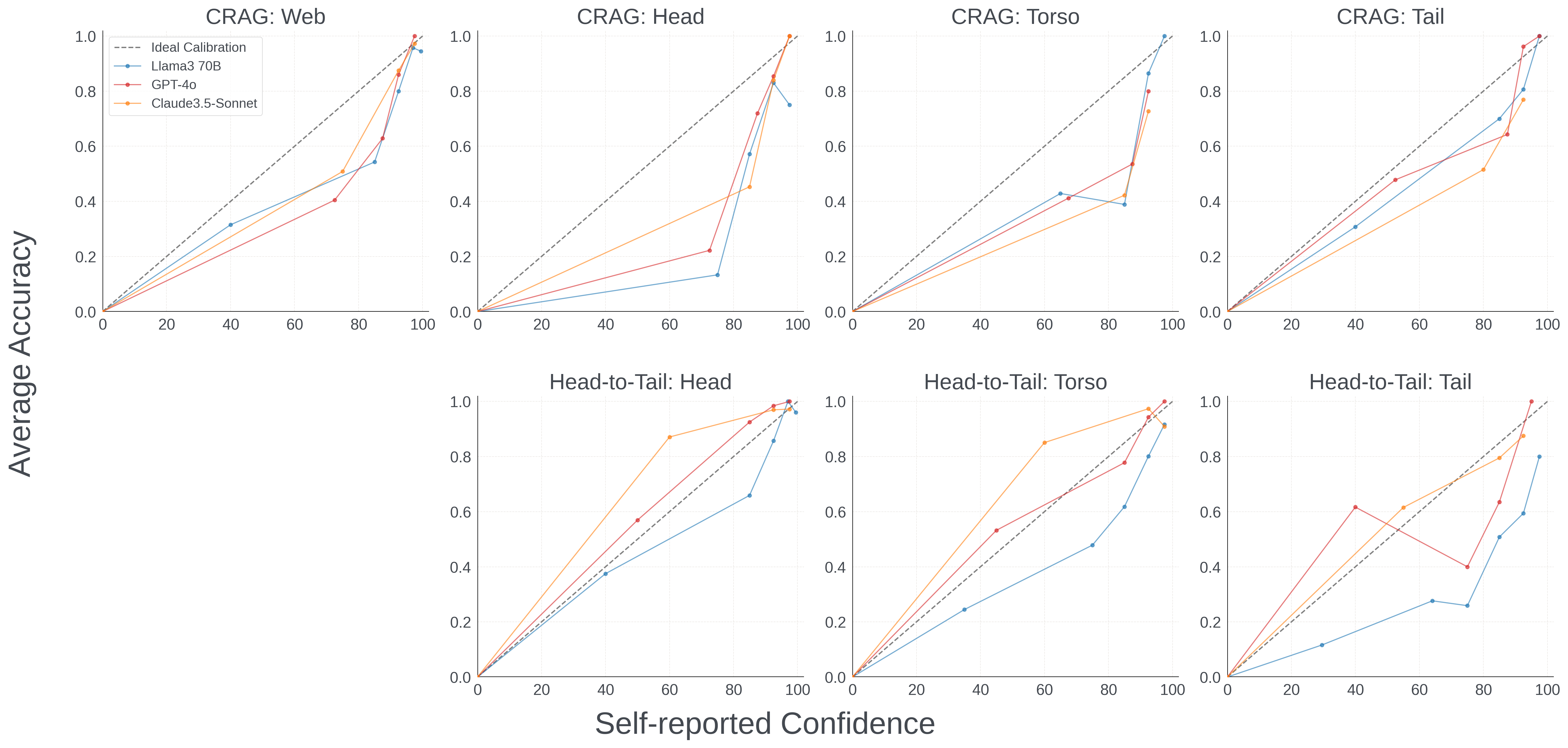}
\caption{Correlation between LLM's self-reported confidences and average accuracies on the CRAG dataset and the Head-to-Tail dataset, categorized by question types.}
\label{fig:confidence-q-type}
\end{figure*}

\paragraph{Influence of entity popularity on confidence:} 
Figure \ref{fig:confidence-q-type} shows the calibration on the Head-to-Tail and CRAG benchmarks, where questions are categorized by entity popularity into \textit{Head, Torso, Tail} (plus \textit{Web} for CRAG). 

Interestingly, we found for simple questions on Head-to-Tail, models are better calibrated for head entities than torso or tail entities (Figure~\ref{fig:confidence-q-type} bottom panels). However, on more complex questions on CRAG, models are better calibrated for tail entities than torso or head entities (Figure~\ref{fig:confidence-q-type} top panels). This shows two different dimensions that can affect the model confidence: entity popularity and question nuances. 

\paragraph{Consistency vs. Accuracy:} 
To measure consistency, we ask LLM the same question 20 times with the temperature set to 1.0, select the most frequent response as the final answer, and calculate its frequency among the 20 times as the consistency score. To be robust against minor differences, we determine the "most frequent" answer based on semantic similarity rather than exact string match.

\begin{figure*}[ht]
\centering
\includegraphics[width=0.9\textwidth]{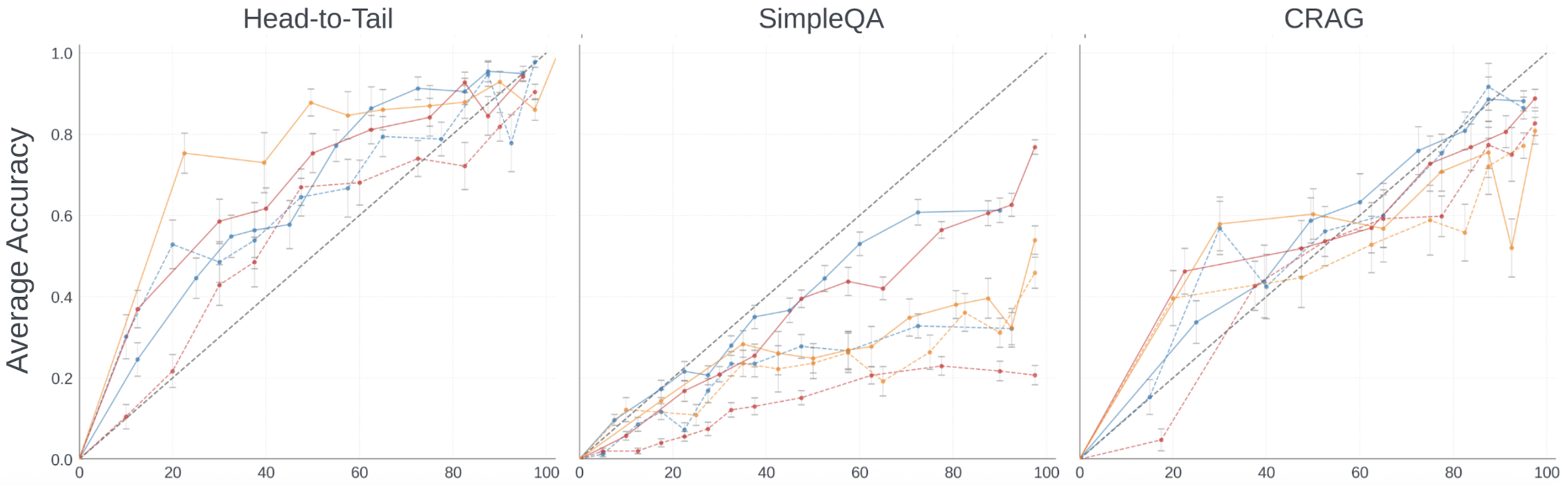}
\caption{LLM's answer consistency is often better calibrated with QA accuracy than self-reported confidences.}
\label{fig:confidence-consistency}
\end{figure*}

Figure~\ref{fig:confidence-consistency} shows that consistency is mostly better calibrated than self-reported confidence, and largely overlays with the ideal calibration on CRAG. In addition, the calibration curve is more linear compared to self-reported confidence in Figure~\ref{fig:confidence-main}.

\section{Prompts}
\label{sec:prompt}

We include a list of prompts used in experiments for reference. Each prompt is cited in the main paper where it is applied. For clarity, we also provide brief descriptions of each prompt here.

\begin{prompt}
\caption{Simple question answer generation prompt}
\label{prompt:short_form_gen_prompt}
\centering
\scriptsize
\begin{tabular}{p{0.48\textwidth}}
\toprule
As Assistant AI, you help answer factual questions. Please keep your responses short and concise and directly provide the answer to the user question without reasoning. Answer only if you are confident; otherwise, respond with ’I am unsure about the answer’. \\
\bottomrule
\end{tabular}
\end{prompt}

Prompt~\ref{prompt:short_form_gen_prompt} is used in several places:
\begin{itemize}
\item {\em Training-data generation}: to prompt Llama-3.1-70B to answer the 3K DBPedia-based questions.
\item {\em SFT}: as the system prompt, unless we state no prompt was used in training.
\item {\em Benchmarks}: as the system prompt at inference, unless we state no prompt was used.
\end{itemize}

\begin{prompt}
\caption{Answer-grading prompt. Used in Section~\ref{subsec:model} to generate training labels: it grades whether the model's answer matches the ground truth. If so, the ground truth becomes the label; otherwise, \textit{“I am unsure about the answer”} is used.
}
\label{prompt:simple_qa_pair_grader}
\centering
\scriptsize
\begin{tabular}{p{0.48\textwidth}}
\toprule
You need to check whether the prediction of a question-answering system
to a question is correct. You should make the judgment based on a list of
ground truth answers provided to you. Your response should be "correct"
if the prediction is correct or "incorrect" if the prediction is wrong.\\\

Example 1:\\
Question: Who authored The Taming of the Shrew (published in 2002)?
Ground truth: ["William Shakespeare", "Roma Gill"]\\
Prediction: W Shakespeare\\
Correctness: correct\\\

Example 2:\\
Question: Who authored The Taming of the Shrew (published in 2002)?
Ground truth: ["William Shakespeare", "Roma Gill"]\\
Prediction: Roma Gill and W Shakespeare\\
Correctness: correct\\\

Example 3:\\
Question: Who authored The Taming of the Shrew (published in 2002)?
Ground truth: ["William Shakespeare", "Roma Gill"]\\
Prediction: Roma Shakespeare\\
Correctness: incorrect\\\

Example 4:\\
Question: What country is Maharashtra Metro Rail Corporation Limited
located in?\\
Ground truth: ["India"]\\
Prediction: Maharashtra\\
Correctness: incorrect\\\

Example 5:\\
Question: What’s the job of Song Kang-ho in Parasite (2019)?
Ground truth: ["actor"]\\
Prediction: He plays the role of Kim Ki-taek, the patriarch of the Kim
family.\\\
Correctness: correct \\
\bottomrule
\end{tabular}
\end{prompt}

Prompt~\ref{prompt:simple_qa_pair_grader} is used in two places:
\begin{itemize}
    \item Training-data generation: to grade whether the model's answer matches the ground truth.
    \item Benchmarks: to grade model responses when evaluating different models.
\end{itemize}

\begin{prompt}
\caption{Self-reported confidence prompt.}
\label{prompt:self-confidence}
\centering
\scriptsize
\begin{tabular}{p{0.48\textwidth}}
\toprule
Please answer the following question truthfully to your best knowledge. Provide a confidence score between 0 to 1 representing your confidence in the answer. 

Question: \{\{\{question\}\}\}

Your response must be in the following JSON format.\\ 
\{``answer'': ``Your answer here'', ``confidence\_score'': float number\}\\
Remember, your answer MUST be a valid json string with two keys: ``answer'', ``confidence\_score''. \\
\bottomrule
\end{tabular}
\end{prompt}
Prompt~\ref{prompt:self-confidence} is used in Appendix~\ref{appen:confidence} to make the model output a confidence score alongside its answer, studying how self-reported confidence and answer consistency each relate to answer accuracy in Appendix~\ref{appen:confidence}.

\begin{prompt}
\caption{Vanilla Prompt from Paper~\cite{ni-etal-2024-llms}.}
\label{prompt:vanilla}
\centering
\scriptsize
\begin{tabular}{p{0.48\textwidth}}
\toprule
Answer the following question based on your internal knowledge with one or few words. If you are sure the answer is accurate and correct, please say "certain" after the answer. If you are not confident with the answer, please say "uncertain".\\
\bottomrule
\end{tabular}
\end{prompt}

\begin{prompt}
\caption{Punish+Explain Prompt from Paper~\cite{ni-etal-2024-llms}.}
\label{prompt:punish_explain}
\centering
\scriptsize
\begin{tabular}{p{0.48\textwidth}}
\toprule
Answer the following question based on your internal knowledge with one or few words and explain why you give this answer. If you are sure the answer is accurate and correct, please say "certain" after the answer. If you are not confident with the answer, please say "uncertain". You will be punished if the answer is not right but you say "certain".\\
\bottomrule
\end{tabular}
\end{prompt}

Prompt~\ref{prompt:vanilla} and~\ref{prompt:punish_explain} are from Paper~\cite{ni-etal-2024-llms}, and are used to generate results for row Prompt and Prompt+ in Table~\ref{tab:overall}.

\section{{\ConfQA} experiment and full ablation study}
We discuss additional details for {\ConfQA} experiment and include the full ablation study as a complementary document to the main paper. 
\subsection{Fine-tuning implementation}
\label{appen:scaling-law}
To determine how many data samples and fine-tuning steps are needed for optimal performance, we run a simple scaling-law study.

We prepared 27K DBPedia question-answer pairs (9k each for head, torso and tail entities) and ran 10K steps on 4 hosts, 8 processes per host, with batch size 1. Around 100 steps gives the best performance, with more causing over-fitting. At this setting, 100 steps is one epoch for 3200 samples. We thus selected 3K high quality instances (1K each for head/torso/tail) and fine-tuned for one epoch. The 1K/1K/1K split follows the real-world distribution of entity popularity by definition (see the Head-to-Tail paper~\cite{headtotail} section 2.1). Head/torso/tail entities are also discussed with the Head-to-Tail benchmark in Appendix~\ref{appen:data}.

\subsection{Full ablation study}
\label{appen:ablation}
We compare {\ConfQA} with more alternatives than in the main content, as shown in Table~\ref{tab:ablation-full}.
\begin{itemize}
    \item {\ConfQA} No-dampener: same training as {\ConfQA}. Does not apply the dampener in inference.
    \item Gen-as-label: the same strategy to choose questions the model can answer correctly as {\ConfQA}, but use model generation as the true label, rather than the ground truth.
    \item IDK: the same as the IDK (DBPedia) in the main paper in Table~\ref{tab:overall}.
    \item No-dampener: the same data labeling as {\ConfQA}, but only use the question answer pairs, excluding the dampener prompt in the training input data. In other words, does not pass dampener as the system prompt.
    \item GT-as-label: feed in the original 3k rows of raw DBPedia data into the fine-tuning, i.e. all labels are ground truth.
    \item Fact-feeding: rather than using only the 3k sample of DBPedia data, also mixing 10k samples from Tulu3 data~\citep{lambert2025tulu3pushingfrontiers}.
    \item R-tuning (DBPedia): using our DBPedia training set and following R-tuning paper to generate labels for SFT.
    \item R-tuning (MMLU): using randomly sampled 3k MMLU samples to generate training set following R-tuning labeling strategy.
    \item MMLU-as-source: the same strategy as {\ConfQA}, but use MMLU as data source. We use the same 3k samples from R-tuning (MMLU).
\end{itemize}

Table~\ref{tab:ablation-full} reports results in two rows. Results on the top are evaluated using no system prompt, i.e. No-dampener applied during inference, only pass in questions to the models. The bottom rows are results with passing the dampener as the system prompt when doing inference.

{\ConfQA} is overall more balanced that improves Missing rate to certain level, without impacting Correctness too much, while reducing Incorrectness to less than 5\%. As shown in main paper Table~\ref{tab:overall}, this leads to the optimal triggering F$_{msr}$.

We are not going to discuss each model one by one, but highlight two observations: (1) as we discussed in Section~\ref{subsec:confqa}, generating training data from MMLU instead of DBPedia yields low hallucination ($<2\%$) but far lower correctness. This is likely because MMLU's diverse tasks reduce the model's overall confidence (Intuition \#3). (2) {\em Fact-feeding} combines our {\ConfQA} fine-tuning data with Tulu facts, drops hallucinations but also correctness, similar to R-tuning. We suspect this is because what our training data teach the LLM (saying unsure) is of different purpose from what the extra Tulu facts teach the LLM (feeding knowledge), when mixed together can offset each other and cause confusion.

\begin{table*}[h]
\centering
\resizebox{1.95\columnwidth}{!}{%
\scriptsize
\begin{tabular}{l|rrrr|rrrr|rrrr|rrrr|}

\toprule
\multirow{2}{*}{\bf Model} & \multicolumn{4}{c|}{\textbf{DBpedia} (in-domain)} & \multicolumn{4}{c|} {\textbf{IMDb} (out-of-domain)} & \multicolumn{4}{c|} {\textbf{SimpleQA} (out-of-domain)}& \multicolumn{4}{c|} {\textbf{CRAG} (out-of-domain)}  \\

\cmidrule(rl){2-5} \cmidrule(rl){6-9} \cmidrule(rl){10-13} \cmidrule(rl){14-17} 

& \bf Corr & \bf Miss & \bf Incor & \bf Fac & \bf Corr & \bf Miss & \bf Incor & \bf Fac & \bf Corr & \bf Miss & \bf Incor & \bf Fac & \bf Corr & \bf Miss & \bf Incor & \bf Fac \\
\midrule

Llama-3.1 Oracle & 47.2 & 21.5 & 31.3 & 15.9 & 41.0 & 33.3 & 25.7 & 15.3 & 16.0 & 42.7 & 41.3 & -25.3 & \textbf{60.3} & 11.7 & 28.0 & 32.3 \\
{\ConfQA} No-dampener & 49.0 & 33.5 & \textit{17.5} & \textbf{31.5} & 43.0 & 42.0 & \textit{16.0} & \textit{27.0} & 17.3 & 55.8 & 26.8& \textit{-9.5}& 57.0&19.6&23.4&33.6 \\

* Gen-as-label & 48.7 & 31.7 & 19.7 & 29.0 &
42.5 & 39.5 & 18.0 & 24.5 &
17.7 & 52.4 & 29.9 & -12.3 &
57.6 & 18.4 & 24.0 & 33.6 \\

* IDK (no-dampener) & 44.5 & 40.3& \textbf{15.2} & \textit{29.3} & 40.7& 45.8& \textbf{13.5} & \textbf{27.2} & 14.4& 65.0& \textbf{20.6} & \textbf{-6.2} & 56.9 &21.5& \textit{21.7} & \textit{35.2}\\
* No-dampener & 42.0 & 34.7 & 23.3 & 18.7 & 
40.2 & 38.0 & 21.8 & 18.4 &
12.0 & 66.4 & \textit{21.6} & -9.6 &
52.6 & 31.2 & \textbf{16.2} & \textbf{36.4}\\
* GT-as-label & 48.7 & \textbf{1.5} & 49.8 & -1.1 & 42.0 & \textbf{0.2} & 57.8 & -15.8 &  18.9 & \textbf{2.7} & 78.5 & -59.6 & 58.1 & \textbf{5.3} & 36.6 & 21.5 \\
* Fact-feeding & 50.0 & 26.8 & 23.2 & 26.8 &
43.3 & 35.5 & 21.2 & 22.1 & 18.1 &
41.2 & 40.7 & -22.6 & 56.9 & 16.5 & 26.6 & 30.3 \\

* R-tuning (DBPedia) & \textbf{53.7} & \textit{6.7} & 39.7 & 14.0 & 44.5 & \textit{11.3} &  44.2 &  0.3 &
 \textbf{22.5} & \textit{13.5} & 64.0 & -41.5 & \textit{58.7} & \textit{8.7} & 32.6 & 26.1 \\
* R-tuning(MMLU) & 50.2 &19.2 &30.7 &19.5 &\textbf{45.5} &28.2 &26.3 & 19.2 &
20.3 & 38.0 & 41.7 & -21.4 & 57.8 & 17.1 & 25.1 &32.7 \\

* MMLU-as-source & \textit{50.5} & 21.8 & 27.7 & 22.8 & \textit{44.2} & 32.7 & 23.2 & 21.0 &
\textit{20.4} & 39.9 & 39.8 & -19.4 & 56.1 & 18.8 & 25.1 & 31.0 \\

\cmidrule(rl){2-5} \cmidrule(rl){6-9} \cmidrule(rl){10-13} \cmidrule(rl){14-17} 
Llama-3.1 Dampen & \textit{47.0} & \textit{26.8} & 26.2& 20.8& \textit{40.7} & \textit{36.2} & 23.2& 17.5
&\textit{16.8} & \textit{48.0} &35.2 &-18.4 &\textbf{57.5} & \textit{22.3} &20.2 & \textbf{37.2}  \\
{\ConfQA} & 31.5 & 63.3 & 5.2 & \textbf{26.3} & 32.5 & 63.3 & 4.2 & \textbf{28.3}
&4.9 & 93.1 & 2.1 & \textbf{2.8} & 39.4 &56.2 & 4.4 &35.0 \\

* Gen-as-label (D) & 28.5 & 65.7 & 5.8 & \textit{22.7} & 
27.7 & 69.8 & 2.5 & \textit{25.2} &
3.1 & 96.0 & 1.9 & \textit{1.2} &
32.7 & 64.0 & 3.3 & 29.4 \\

* IDK & 17.0 & 81.5 & \textbf{1.5} &15.5&22.0 &77.0 & \textbf{1.0} &21.0 &0.6 & 99.1 & \textbf{0.2} & 0.4 & 20.7 &78.2 & \textit{1.1} &19.6 \\
* No-dampener (D) & 36.0& 50.2& 13.8& 22.2& 
34.2& 50.7& 15.2& 19.0& 
5.8& 87.5& 6.7& -0.9& 
46.0& 44.2& 9.8& \textit{36.2}\\
* GT-as-label (D) & \textbf{48.0} & \textbf{2.8} & 49.2 & -1.2 & \textbf{41.2} &  \textbf{4.3} &  54.5 &  -13.3 & \textbf{17.8} & \textbf{13.7} & 68.4 & -50.6 & \textit{53.7} & \textbf{14.2} &  32.1 & 21.6 \\

* Fact-feeding (D)& 20.7& 76.7& 2.7 & 18.0 & 
25.5& 70.7& 3.8& 21.7 &
2.5 &  94.7 &  2.8 & -0.3 &
22.4 &  74.5 & 3.1 & 19.3\\

* R-tuning (DBPedia) & 24.5 & 67.8 & 7.7 &16.8 &25.3 & 70.2 & 4.5 & 20.8 &
3.7 &83.3 &13.0 &-9.3 & 31.6 & 55.0 &13.4 &18.2 \\
* R-tuning(MMLU) &24.3 &67.3 &8.3 &16.0 &28.2 &60.5 &11.3 &16.9 &
5.8 & 85.1 & 9.1 &-3.3 & 31.3 &56.5 &12.1 &19.2 \\

* MMLU-as-source (D) & 8.2 & 89.8 & \textit{2.0} &6.2 & 16.7 & 82.0 & \textit{1.3} & 15.4 &
0.6 & 98.8 & \textit{0.5} & 0.1 & 7.0 & 92.7 & \textbf{0.3} &6.7 \\
\bottomrule
\end{tabular}
}
\caption{Ablation study of {\ConfQA}, showing effectiveness of our fine-tuned model to reduce hallucination and improve factuality (all numbers are in percentage~(\%); bold indicates the best
performance and italic the second best.)}
\label{tab:ablation-full}
\end{table*}

\subsection{Answer distributions for entity popularity}
\begin{figure*}[h]
  \centering
     \begin{subfigure}[b]{0.31\textwidth}
         \centering
         \includegraphics[width=\textwidth]{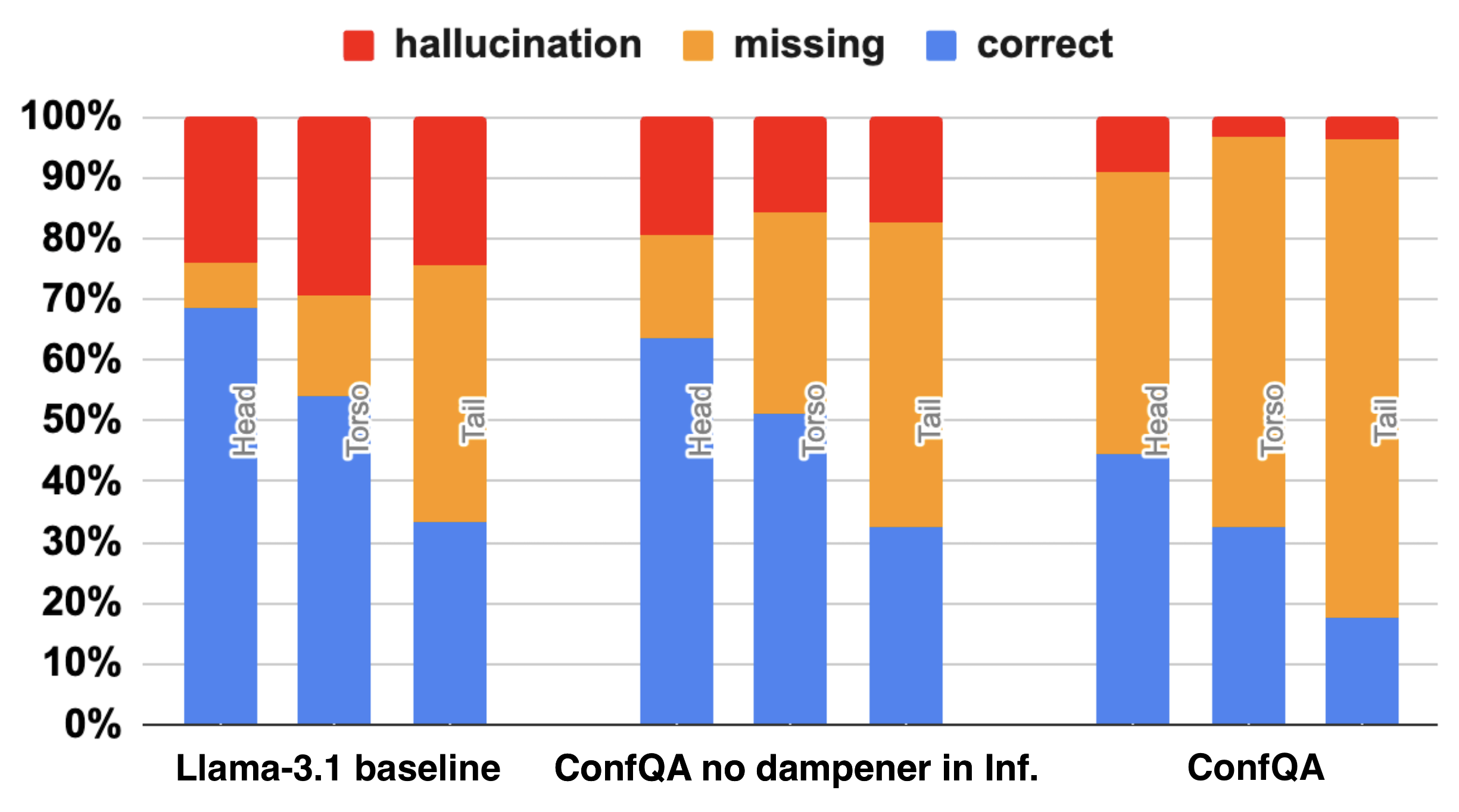}
         \caption{DBPedia}
         \label{fig:popularity_dbpedia}
     \end{subfigure}
     \hfill
     \begin{subfigure}[b]{0.31\textwidth}
         \centering
         \includegraphics[width=\textwidth]{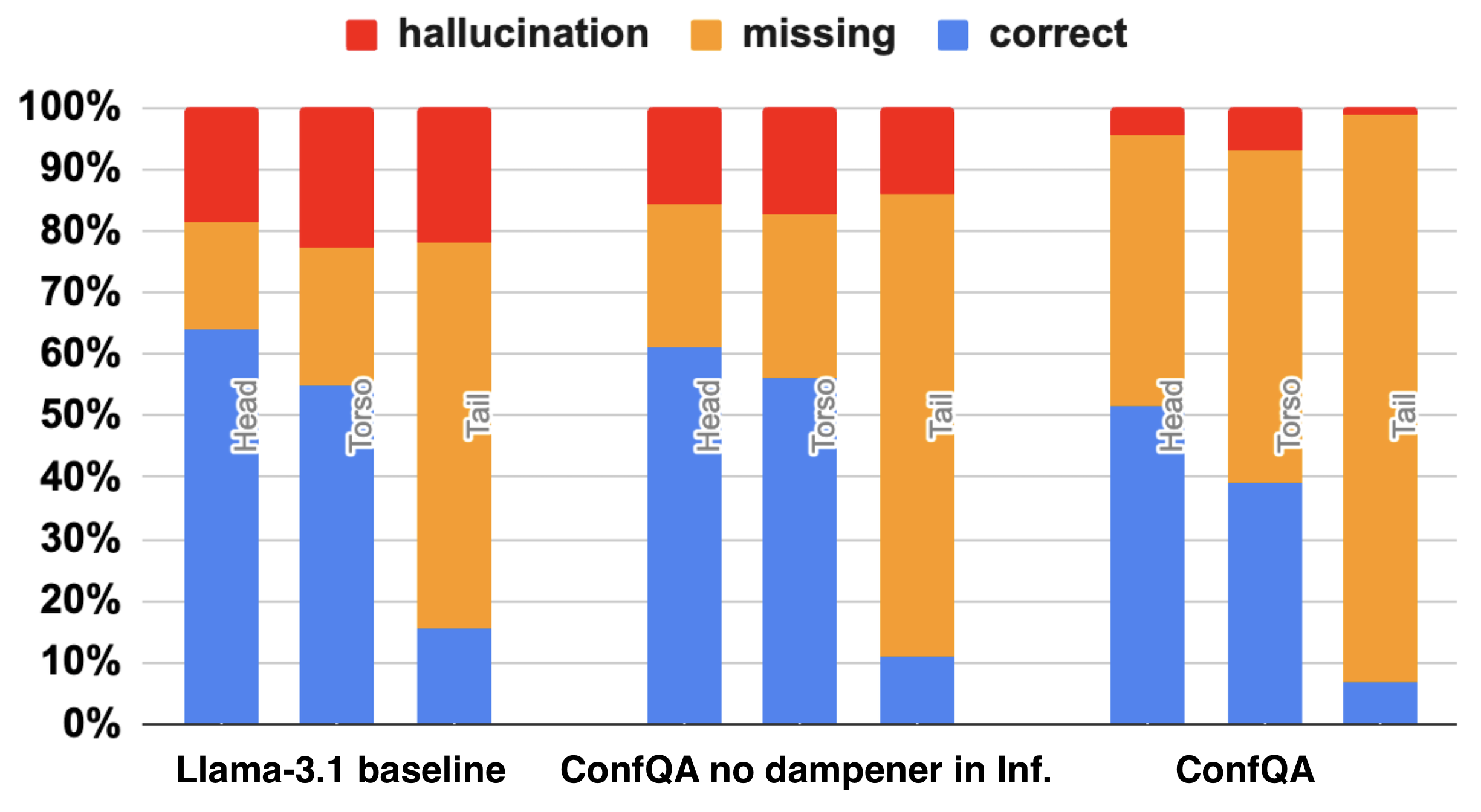}
         \caption{IMDb}
         \label{fig:popularity_imdb}
     \end{subfigure}
     \hfill
     \begin{subfigure}[b]{0.31\textwidth}
         \centering
         \includegraphics[width=\textwidth]{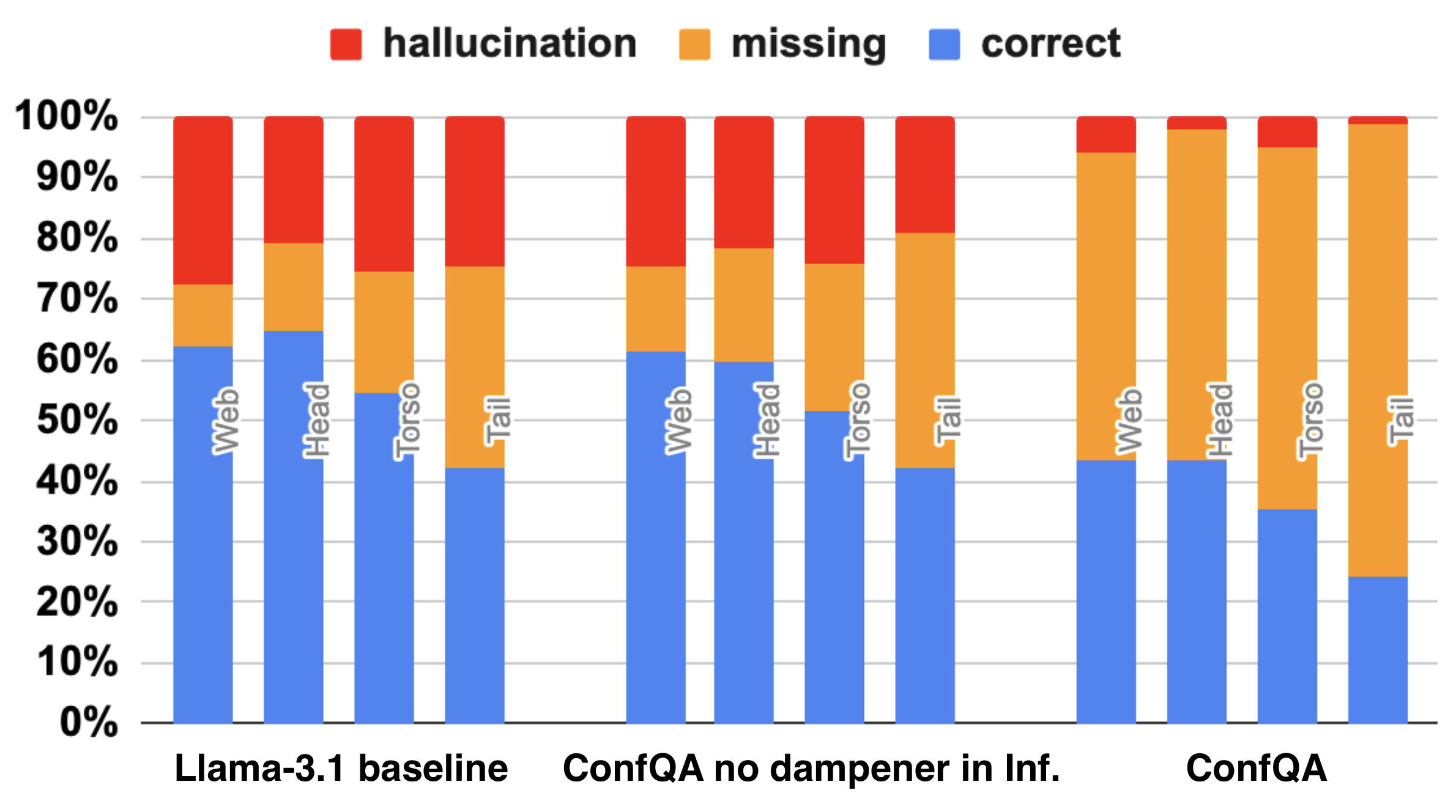}
         \caption{CRAG}
         \label{fig:popularity_crag}
     \end{subfigure}
  \caption{{\ConfQA} suppresses more (i.e., gives more missing answers) on less popular entities; that is, missing rate for Tail is larger than Torso, which is in turn larger than Head.}
  \label{fig:mainpopularity}
\end{figure*}
Figure~\ref{fig:mainpopularity} shows the distribution of correct, missing, and incorrect answers for entities of different popularity (Head, Torso, Tail), before and after fine-tuning, with and without dampening. It confirms that fine-tuning suppresses hallucinations (Intuition \#1), and the dampener prompt further reduces hallucinations (Intuition \#2). Additionally, it shows {\ConfQA} suppresses more on long-tail facts, where it lacks confidence.

\subsection{p-Value of {\ConfQA} models}
\label{sec:p_value}
We compute p-Values for {\ConfQA} model on the hallucination reduction metrics comparing with baseline Llama-3.1 Oracle and Llama3.1 Dampen, and report in Table~\ref{tab:overall_p_value}. We compare {\ConfQA} with Llama3.1 Dampen and {\ConfQA} No-dampener with Llama-3.1 Oracle. We use McNemar p-Value since this is a paired test. For simplicity, we assume at least 70\% of the questions are the same in baseline vs {\ConfQA}. For factuality, we compute independent z-test p-Value. The results show that the improvements on Hallucination reduction shown in Table~\ref{tab:overall} are statistically significant on all benchmarks. Factuality improvements are also statistically significant for DBPedia, IMDb and SimpleQA benchmarks.

\begin{table*}[h!]
\centering
\scriptsize
\begin{tabular}{l|rr|rr|}

\toprule
\bf Model & \bf Incor (p-Value) & \bf Fac (p-Value) & \bf Incor (p-Value) & \bf Fac (p-Value)  \\
\midrule
 & \multicolumn{2}{c|}{\textbf{DBPedia} (in-domain)} & \multicolumn{2}{c|} {\textbf{IMDb} (out-of-domain)} \\
\cmidrule(rl){2-3} \cmidrule(rl){4-5} 

Llama-3.1 Oracle & 31.3 & 15.9 & 25.7 & 15.3  \\
Llama-3.1 Dampen  & 26.2& 20.8& 23.2& 17.5 \\
\cmidrule(rl){2-3} \cmidrule(rl){4-5} 
{\ConfQA} No-dampener &17.5 (5.5E-12) & 31.5 (2.1E-10) &
16.0 (7.2E-08) & 27.0 (7.0E-07)  \\
{\ConfQA}  & 5.2 (<1.0E-16) &26.3 (2.5E-02) & 
4.2 (<1.0E-16) & 28.3 (8.5E-06) \\

\toprule
 & \multicolumn{2}{c} {\textbf{SimpleQA} (out-of-domain)} & \multicolumn{2}{c}{\textbf{CRAG} (out-of-domain)}  \\

\cmidrule(rl){2-3} \cmidrule(rl){4-5} 

Llama3.1 Oracle & 41.3 & -25.3& 28.0 & 32.3 \\
Llama3.1 Dampen  &35.2&-18.4& 20.2& 37.2 \\
\cmidrule(rl){2-3} \cmidrule(rl){4-5}  
{\ConfQA} No-dampener & 26.8 (<1.0E-16) & -9.5 (<1.0E-16) & 
 23.4 (6.2E-03) & 33.6 (6.2E-01) \\
{\ConfQA} & 2.1 (<1.0E-16) & 2.8 (<1.0E-16) & 
4.4 (<1.0E-16) &35.0 (4.1E-01) \\
\bottomrule
\end{tabular}
\caption{Factuality improvement on short-form benchmarks with p-Value; {\ConfQA} models can reduce hallucination (Incor) to less than 5\% with the dampener prompt with significant difference. All numbers are in percentage~(\%).}
\label{tab:overall_p_value}
\end{table*}

\subsection{Gemma and QWen SFT}
\label{appen:other_model}

We conducted similar experiments on the QWen2.5-7B-Instruct model~\citep{qwen2025qwen25technicalreport} and the Gemma-3-4B-IT model~\citep{gemmateam2025gemma3technicalreport} through the Huggingface framework. Training data is generated using the logic described in Section~\ref{subsec:model}:
\begin{itemize}
    \item The same 3k of DBPedia simple question answer data, and labeled using the logic in~\ref{subsec:model}.
    \item Evaluations are done using the same 4 short-form generation data sets.
\end{itemize}
Train using NVIDIA A100 80G 8 GPU. For both fine-tuning, we run one epoch, with learning rate 2e-4 and gradient accumulation steps 2.

\begin{table*}[t!]
\centering
\resizebox{1.5\columnwidth}{!}{%

\begin{tabular}{l|rrrr|rrrr|}

\toprule
\bf Model & \bf Corr & \bf Miss & \bf Incor & \bf Fac. & \bf Corr & \bf Miss & \bf Incor & \bf Fact.   \\
\midrule
 & \multicolumn{4}{c|}{\textbf{DBPedia} (in-domain)} & \multicolumn{4}{c|} {\textbf{IMDb} (out-of-domain)} \\

\cmidrule(rl){2-5} \cmidrule(rl){6-9} 

QWen2.5 & 21.3 & \textbf{25.8} & 52.8 & -31.5 & 19.3 & \textbf{15.7} & 65.0 & -45.7 \\
QWen2.5 Dampen & 12.0 & 64.2 & 23.5 & -11.5  & 14.7 & 56.7 & 28.3 & -13.6 \\

\cmidrule(rl){2-5} \cmidrule(rl){6-9} 
{\ConfQA} No-dampener in Inf. & \textbf{21.5} & 50.8 & 27.7 & -6.2  & \textbf{19.5} & 31.5 & 49.0 & -29.5 \\
{\ConfQA} & 8.3 & 88.7 & \textbf{3.0} & \textbf{5.3} & 12.7 & 74.0 & \textbf{13.3} & \textbf{-0.6} \\

\toprule
 & \multicolumn{4}{c} {\textbf{SimpleQA} (out-of-domain)} & \multicolumn{4}{c}{\textbf{CRAG} (out-of-domain)}  \\

\cmidrule(rl){2-5} \cmidrule(rl){6-9} 
QWen2.5 & \textbf{3.8} & \textbf{21.6} & 74.5 & -70.7 & 22.7 & \textbf{30.1} & 47.2 & -24.5 \\
QWen2.5 Dampen  & 1.7 & 71.5 & 26.8 & -25.1 & 15.7 & 57.8 & 26.5 & -10.8 \\

\cmidrule(rl){2-5} \cmidrule(rl){6-9} 
{\ConfQA} No-dampener in Inf. & 2.8 & 61.0 & 36.2 & -33.4 & \textbf{24.3} & 41.1 & 34.6 & -10.3 \\
{\ConfQA} & 0.8 & 86.1 & \textbf{13.1} & \textbf{-12.3} & 8.3 & 85.2 & \textbf{6.5} & \textbf{1.8}  \\
\bottomrule

\end{tabular}
}
\caption{Overall factuality improvement on short-form benchmarks for QWen2.5-7B-Instruct; {\ConfQA} can reduce hallucination to around  10\% with the dampener prompt. All numbers are in percentage~(\%).}
\label{tab:qwen_overall}
\end{table*}

\begin{table*}[t!]
\centering
\resizebox{1.5\columnwidth}{!}{%

\begin{tabular}{l|rrrr|rrrr|}

\toprule
\bf Model & \bf Corr & \bf Miss & \bf Incor & \bf Fac.  & \bf Corr & \bf Miss & \bf Incor & \bf Fact.  \\
\midrule
 & \multicolumn{4}{c|}{\textbf{DBPedia} (in-domain)} & \multicolumn{4}{c|} {\textbf{IMDb} (out-of-domain)} \\
\cmidrule(rl){2-5} \cmidrule(rl){6-9} 

Gemma3 & 19.5 & 4.5 & 76.0 & -56.5  & 17.2 & 2.0 & 80.8 & -63.6  \\
Gemma3 Dampen & 20.5 & \textbf{4.3} & 75.2 & -54.7 & \textbf{19.5} & \textbf{0.3} & 80.2 & -60.7  \\
\cmidrule(rl){2-5} \cmidrule(rl){6-9} 

{\ConfQA} No-dampener in Inf.& \textbf{22.3} & 38.8 & 38.8 & -16.5 & 16.2 & 24.8 & 59.0 & -42.8  \\
{\ConfQA} & 6.0 & 92.2 & \textbf{1.8} & \textbf{4.2} & 9.0 & 85.2 & \textbf{5.8} & \textbf{3.2} \\
\toprule
 & \multicolumn{4}{c} {\textbf{SimpleQA} (out-of-domain)} & \multicolumn{4}{c}{\textbf{CRAG} (out-of-domain)}  \\
\cmidrule(rl){2-5} \cmidrule(rl){6-9} 

Gemma3 & \textbf{4.0} & 1.8 & 94.2 & -90.2 & 26.8 & 3.0 & 70.2 & -43.4 \\
Gemma3 Dampen & 3.7 & \textbf{0.9} & 95.4 & -91.7 & \textbf{29.1} & \textbf{2.6} & 68.2 & -39.1 \\
\cmidrule(rl){2-5} \cmidrule(rl){6-9} 

{\ConfQA} No-dampener in Inf. & 2.5 & 43.1 & 54.5 & -52.0 & 19.3 & 39.3 & 41.4 & -22.1 \\
{\ConfQA} & 0.2 & 97.2 & \textbf{2.5} & \textbf{-2.3} & 3.3 & 95.0 & \textbf{1.7} & \textbf{1.6} \\
\bottomrule

\end{tabular}
}
\caption{Overall factuality improvement on short-form benchmarks for Gemma-3-4B-IT; {\ConfQA} can reduce hallucination to below 5\% with the dampener prompt. All numbers are in percentage~(\%).}
\label{tab:gemma_overall}
\end{table*}

Table~\ref{tab:qwen_overall} shows experiments results using the QWen2.5-7B-Instruct model as baseline model. Table~\ref{tab:gemma_overall} shows experiment results from fine-tuning Gemma-3-4B-IT. A few observations based on these results:
\begin{itemize}
    \item Our proposed method {\ConfQA} can reduce hallucination (Incor) by 13-50\% for QWen 7B when applying the dampener prompt.
    \item The hallucination could be reduced to close or below 5\% for Gemma 4B model in the same case.
    \item Qwen 7B has fairly low correctness, and {\ConfQA} can further reduce it as it changes low-confidence answers into unsure answers. However, the factuality increases by 12-37\%, showing that it reduces much more hallucinations than correct answers.
    \item Similar for Gemma model: fairly low correctness, as it is a even smaller model. Comparing to QWen, the factuality increase is more effective for Gemma model: increases by 20-89\%.
    \item Transferability: the fine-tuning on DBPedia atomic question answering pairs could extend to out-of-domain datasets.
\end{itemize}

\subsection{Dampener prompt wording}
\label{appen:prompt_wording}

To understand how sensitive is the approach to the exact wording of the dampening prompt, we evaluated the fine-tuned {\ConfQA} model with different wordings in the prompt. The {\ConfQA} prompt is listed in Appendix Prompt~\ref{prompt:short_form_gen_prompt}. In this experiment, we only change the dampener part of the whole prompt, i.e. {\emph{Answer only if you are confident; otherwise, respond with ’I am unsure about the answer’.}}
In the experiment, Prompt 1: {\emph{Answer only if you are confident; otherwise, respond with 'I am unsure'}}, Prompt 2: {\emph{Answer if you are confident; otherwise, respond with 'I am unsure about the answer'}}.

Table~\ref{tab:prompt_wording} shows results from using different prompts in the evaluation. We see that Prompt 1 and 2 give very similar results compared to baseline {\ConfQA}.

\begin{table*}[t]
\centering
\resizebox{1.3\columnwidth}{!}{%

\begin{tabular}{l|rrrr|rrrr|}

\toprule
\bf Prompt & \bf Corr & \bf Miss & \bf Incor & \bf Fac.  & \bf Corr & \bf Miss & \bf Incor & \bf Fact.  \\
\midrule
 & \multicolumn{4}{c|}{\textbf{DBPedia} (in-domain)} & \multicolumn{4}{c|} {\textbf{IMDb} (out-of-domain)} \\
\cmidrule(rl){2-5} \cmidrule(rl){6-9} 

{\ConfQA} & 31.5 & 63.3 & 5.2 & 26.3  & 32.5 & 63.3 & 4.2 & 28.3  \\
\cmidrule(rl){2-5} \cmidrule(rl){6-9} 
Prompt 1 & 30.7 & 64.0 & 5.3 & 25.4 & 31.2 & 65.2 & 3.7 &  27.5 \\
Prompt 2 & 32.0 & 63.5 & 4.5 & 27.5 & 32.0 & 64.7 & 3.3 & 28.7  \\
\toprule
 & \multicolumn{4}{c} {\textbf{SimpleQA} (out-of-domain)} & \multicolumn{4}{c}{\textbf{CRAG} (out-of-domain)}  \\
\cmidrule(rl){2-5} \cmidrule(rl){6-9} 

{\ConfQA} & 4.9 & 93.1 & 2.1 & 2.8 & 39.4 & 56.2 & 4.4 & 35.0 \\
\cmidrule(rl){2-5} \cmidrule(rl){6-9} 
Prompt 1 & 5.0 & 93.3 & 1.7 &  3.3 & 39.9 & 55.9 & 4.2 & 35.7  \\
Prompt 2 & 5.9 & 92.7 & 1.3 &  4.6 & 37.9 & 58.3 & 3.9 & 34.0  \\
\bottomrule

\end{tabular}
}
\caption{Overall factuality difference with different wording in the dampener prompt. All numbers are in percentage~(\%).}
\label{tab:prompt_wording}
\end{table*}

\section{Dynamic questions identification}
\label{appen:dynamic}
In Section~\ref{subsec:trigger_rag}, we briefly mentioned that dynamic questions--those requiring up-to-date information--should be answered by the RAG pipeline. In real application, we fine-tune a small-sized model (such as Llama-3.1 8B Instruct) to do this classification task: identify if a question is dynamic (whose answer may change over time) or static (whose answer is static).

\paragraph{Training data and prompt:} Take CRAG training data as example, instead of using ground truth answer as label, we label the questions as either "dynamic" or "static". We use prompt to instruct the model to output only "dynamic" or "static". Respond with "dynamic" if the answer to the question could change over time, and respond with "static" if the answer to the question does not change over time.
The same supervised fine-tuning is applied to the model. We normally observe 80-90\% precision and recall of this method in identifying dynamic questions. 
We omit additional details as this is not relevant to the key content of this paper.

\end{document}